\documentclass{article}

\usepackage[preprint]{neurips_data_2022}
\setcitestyle{authoryear, open={(},close={)}}

\usepackage[utf8]{inputenc} % allow utf-8 input
\usepackage[T1]{fontenc}    % use 8-bit T1 fonts
\usepackage{hyperref}       % hyperlinks
\usepackage{url}            % simple URL typesetting
\usepackage{booktabs}       % professional-quality tables
\usepackage{amsfonts}       % blackboard math symbols
\usepackage{nicefrac}       % compact symbols for 1/2, etc.
\usepackage{microtype}      % microtypography
\usepackage{xcolor}         % colors
\usepackage{graphicx}
\usepackage{soul}
\usepackage{enumitem}
\usepackage{bm}
\usepackage{wrapfig}
\usepackage{lipsum}
\usepackage{multirow}

\usepackage{pifont}

\title{K-Radar: 4D Radar Object Detection for Autonomous Driving in Various Weather Conditions}
\author{%
  Dong-Hee Paek$^{1}$\thanks{co-first authors} \quad\quad Seung-Hyun Kong$^{1}$\footnotemark[1] \thanks{corresponding author} \quad\quad Kevin Tirta Wijaya$^{2}$\\
  $^1$CCS Graduate School of Mobility \quad\quad $^{2}$Robotics Program\\
  KAIST\\
  \texttt{\{donghee.paek, skong, kevin.tirta\}@kaist.ac.kr} \\
}

\begin{document}

\maketitle

\begin{abstract}
  Unlike RGB cameras that use visible light bands (384$\sim$769 THz) and Lidars that use infrared bands (361$\sim$331 THz), Radars use relatively longer wavelength radio bands (77$\sim$81 GHz), resulting in robust measurements in adverse weathers.
  Unfortunately, existing Radar datasets only contain a relatively small number of samples compared to the existing camera and Lidar datasets.
  This may hinder the development of sophisticated data-driven deep learning techniques for Radar-based perception.
  Moreover, most of the existing Radar datasets only provide 3D Radar tensor (3DRT) data that contain power measurements along the Doppler, range, and azimuth dimensions.
  As there is no elevation information, it is challenging to estimate the 3D bounding box of an object from 3DRT.
  In this work, we introduce KAIST-Radar (K-Radar), a novel large-scale object detection dataset and benchmark that contains 35K frames of 4D Radar tensor (4DRT) data with power measurements along the Doppler, range, azimuth, and elevation dimensions, together with carefully annotated 3D bounding box labels of objects on the roads.
  K-Radar includes challenging driving conditions such as adverse weathers (fog, rain, and snow) on various road structures (urban, suburban roads, alleyways, and highways). 
  In addition to the 4DRT, we provide auxiliary measurements from carefully calibrated high-resolution Lidars, surround stereo cameras, and RTK-GPS.
  {We also provide 4DRT-based object detection baseline neural networks (baseline NNs) and show that the height information is crucial for 3D object detection.}
%  \hl{We also provide 4DRT-based object detection detection baseline neural networks (baseline NNs) and show that additional height information can significantly improves the 3D detection performance.}
  {And by comparing the baseline NN with a similarly-structured Lidar-based neural network, we demonstrate that 4D Radar is a more robust sensor for adverse weather conditions.}
%  \hl{And by comparing the baseline NN with the representative Lidar-based neural network, we demonstrate that 4D Radar is a more robust sensor under adverse weather conditions.}
%   To demonstrate the robustness of 4D Radar-based perception, we present a 4DRT-based 3D object detection baseline neural network that surpasses Lidar-based 3D object detection network under adverse weather conditions.
%   In addition, we show that the additional elevation information in 4DRT can significantly improves the detection performance. 
  All codes are available at \url{https://github.com/kaist-avelab/k-radar}.
\end{abstract}

\section{Introduction}

An autonomous driving system generally consists of sequential modules of perception, planning, and control.
As the planning and control modules rely on the output of the perception module, it is crucial for the perception module to be robust even under adverse driving conditions.

Recently, various works have proposed deep learning-based autonomous driving perception modules that demonstrate remarkable performances in lane detection \citep{klane, condlane}, object detection \citep{scaledyolo, pointpillars, highway_radar}, and other tasks \citep{vitdepth, deep_slam}.
These works often use RGB images as the inputs to the neural networks due to the availability of numerous public large-scale datasets for camera-based perception.
Moreover, an RGB image has a relatively simple data structure, where the data dimensionality is relatively low and neighboring pixels often have high correlation. 
Such a simplicity enables deep neural networks to learn the underlying representations of images and recognize objects on the image. 

Unfortunately, camera is prone to poor illumination, can easily be obscured by raindrops and snowflakes, and cannot preserve depth information that is crucial for accurate 3D scene understanding of the environment.
On the other hand, Lidar actively emits measuring signals in the infrared spectrum, therefore, the measurements are hardly affected by illumination conditions.
Lidar can also provide accurate depth measurements within centimeters resolution.
%Unfortunately, camera is prone to poor illumination such as low light conditions and glares.
%Camera measurements can also be easily obscured by raindrops and snowflakes.
%In addition, since depth information is not preserved, it is challenging to obtain accurate 3D spatial information of the surrounding environment.
%Compared to camera that only passively measures waves in the visible spectrum, Lidar actively emits infrared waves and measures the reflections, therefore, the measurements are hardly affected by illumination conditions.
%Lidar can also provide depth measurements with centimeters resolution so that the 3D spatial information can be preserved.
%Lidar can also provide accurate depth information within centimeters resolution so that the 3D spatial information can be preserved.
However, Lidar measurements are still affected by adverse weathers since the wavelength of the signals ($\lambda$=850nm$\sim$1550nm) is not long enough to pass through raindrops or snowflakes \citep{lidar_snow}.

% \begin{figure}[tb!]
%   \centering
%   \includegraphics[width=1.0\columnwidth]{fig1.png}
%   \caption{An overview of the signal processing of the FMCW Radar and a visualization of the two main data types (i.e., Radar tensor (RT) and Radar point cloud (RPC)). The RT is a dense data matrix with power measurements in all element along the dimensions through a Fast Fourier Transform (FFT) operation applied to FMCW signals. Since all elements are non-zero values, the RT provides dense information regarding the environment with minimal loss, at a cost of high memory requirement. On the other hand, the RPC is a data type in which target (i.e., object candidate group) information is extracted in the form of a point cloud with a small amount of memory by applying Constant False Alarm Rate (CFAR) algorithm to the RT. Due to the ease of implementing FFT and CFAR directly on the hardware, many Radar sensors provide RPCs as output. However, the RPC may lose a significant amount of information regarding the environment due to the CFAR algorithm.}
%   \label{radar_system}
% \end{figure}

\begin{wrapfigure}{L}{0.68\columnwidth}
  \centering
  \includegraphics[width=0.68\columnwidth]{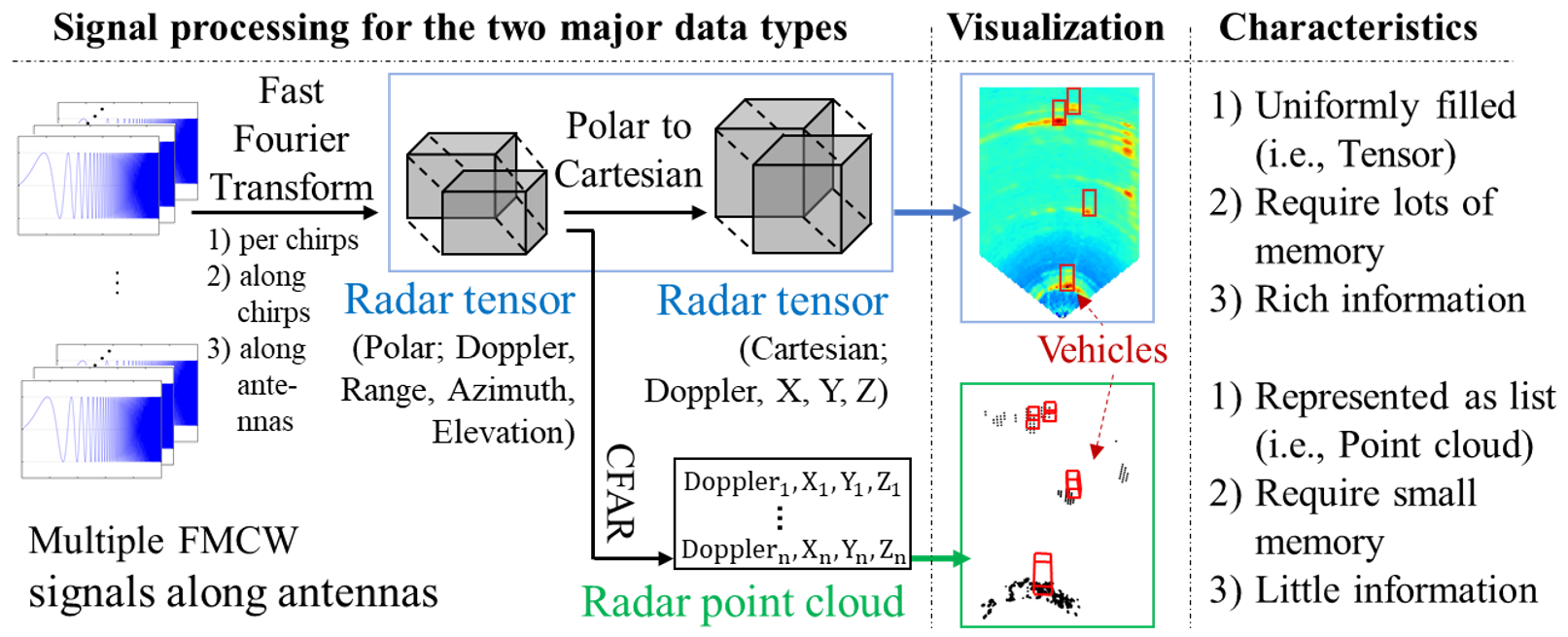}
  \caption{An overview of the signal processing of the FMCW Radar and a visualization of the two main data types (i.e., Radar tensor (RT) and Radar point cloud (RPC)). The RT is a dense data matrix with power measurements in all element along the dimensions through a Fast Fourier Transform (FFT) operation applied to FMCW signals. Since all elements are non-zero values, the RT provides dense information regarding the environment with minimal loss, at a cost of high memory requirement. On the other hand, the RPC is a data type in which target (i.e., object candidate group) information is extracted in the form of a point cloud with a small amount of memory by applying Constant False Alarm Rate (CFAR) algorithm to the RT. Due to the ease of implementing FFT and CFAR directly on the hardware, many Radar sensors provide RPCs as output. However, the RPC may lose a significant amount of information regarding the environment due to the CFAR algorithm.}
  \label{radar_system}
\end{wrapfigure}

Similar to Lidar, a Radar sensor actively emits waves and measures the reflection. 
However, Radar emits radio waves ($\lambda\approx$ 4mm) that can pass through raindrops and snowflakes.
As a result, Radar measurements are robust to both poor illumination and adverse weather conditions.
This robustness is demonstrated in \citep{radar_deep_learning_review}, where a Frequency Modulated Continuous Wave (FMCW) Radar-based perception module is shown to be accurate even in adverse weather conditions and can be easily implemented directly on the hardware.
%The excellent performance of Radar for autonomous driving perception module is demonstrated in \citep{automotive_rdr}, where the Frequency Modulated Continuous Wave (FMCW) Radar-based perception module is shown to be robust under adverse weather conditions and can be easily implemented directly on the hardware.

As FMCW Radars with dense Radar tensor (RT) outputs become readily available, numerous works \citep{prob_radar,zendar,radiate} propose RT-based object detection networks with comparable detection performance to camera and Lidar-based object detection networks.
However, these works are limited to 2D bird-eye-view (BEV) object detection, since FMCW Radars utilized in existing works only provide 3D Radar tensor (3DRT) with power measurements along the Doppler, range, and azimuth dimensions.

In this work, we introduce KAIST-Radar (K-Radar), a novel 4D Radar tensor (4DRT)-based 3D object detection dataset and benchmark. Unlike the conventional 3DRT, 4DRT contains power measurements along the Doppler, range, azimuth, and elevation dimensions so that the 3D spatial information can be preserved, which could enable accurate 3D perception such as 3D object detection with Lidar. To the best of our knowledge, K-Radar is the first large-scale 4DRT-based dataset and benchmark, with 35k frames collected from various road structures (e.g. urban, suburban, highways), time (e.g. day, night), and weather conditions (e.g. clear, fog, rain, snow). In addition to the 4DRT, K-Radar also provides high-resolution Lidar point clouds (LPCs), surround RGB images from four stereo cameras, and RTK-GPS and IMU data of the ego-vehicle.

Since the 4DRT high-dimensional representation is unintuitive to human, we leverage the high-resolution LPC so that the annotators can accurately label the 3D bounding boxes of objects on the road in the visualized point clouds. 
The 3D bounding boxes can be easily transformed from the Lidar to the Radar coordinate frame since we provide both spatial and temporal calibration parameters to correct offsets due to the separations of the sensors and the asynchronous measurements, respectively. 
K-Radar also provides a unique tracking ID for each annotated object that is useful for tracking an object along a sequence of frames.
Examples of information regarding tracking are shown in Appendix I.7.

\begin{figure}[tb!]
  \centering
  \includegraphics[width=1.0\columnwidth]{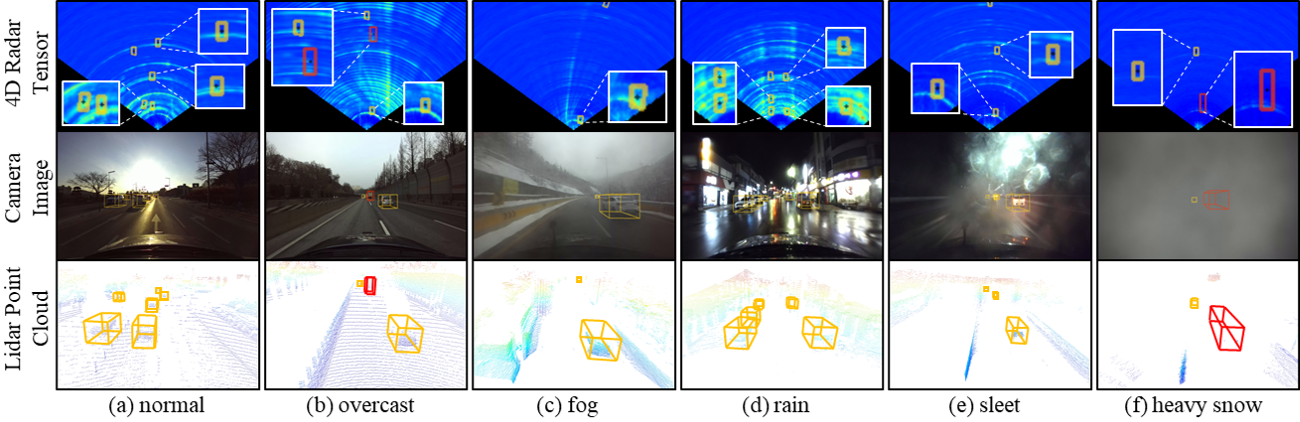}
  \caption{Samples of K-Radar datasets for various weather conditions. Each column shows (1) 4DRTs, (2) front view camera images, and (3) Lidar point clouds (LPCs) of different weather conditions. 4DRTs are represented in a two-dimensional (BEV) Cartesian coordinate system using a series of visualization processes that are described in Section 3.3. In this example, yellow and red bounding boxes represent the sedan and bus or truck classes, respectively. Appendix A contains further samples of K-Radar datasets for each weather condition.}
   % \caption{Samples of K-Radar datasets for various road and weather conditions. (1) 4DRTs, (2) front view camera images, and (3) Lidar point clouds (LPCs) of two different road conditions with the same weather condition are depicted in the two boxes in each column. 4DRTs are represented in a two-dimensional (BEV) Cartesian coordinate system using a series of visualization processes that are described in Section 3.3. In this example, the bounding box is denoted by color-coding the class of the object (sedan: yellow, bus or truck: red). Appendix A contains further samples of K-Radar datasets for each weather condition.}
  \label{kradar}
\end{figure}

To demonstrate the necessity of 4DRT-based perception module, we present a 3D object detection baseline neural network (baseline NN) that directly consumes 4DRT as an input. 
From the experimental results on K-Radar, we observe that the 4DRT-based baseline NN outperforms the Lidar-based network in the 3D object detection task, especially in adverse weather conditions. 
We also show that the 4DRT-based baseline NN utilizing height information significantly outperforms network that only utilizes BEV information. 
Additionally, we publish the complete development kits (devkits) that include: (1) training / evaluation codes for 4DRT-based neural networks, (2) labeling / calibration tools, and (3) visualization tools to accelerate research in the field of 4DRT-based perception.
%Additionally, we publish the complete development kits (devkits) at https://github.com/kaist\_avelab/k-Radar. 
%The devkits include: (1) training / evaluation code for 4DRT-based neural networks, (2) labeling / calibration tools, and (3) visualization tools to accelerate research in the field of 4DRT-based perception for autonomous driving.

In a summary, our contributions are as follow,
\begin{itemize}[noitemsep]
    \item We present a novel 4DRT-based dataset and benchmark, K-Radar, for 3D object detection. 
    To the best of our knowledge, K-Radar is the first large-scale 4DRT-based dataset and benchmark with diverse and challenging illumination, time, and weather conditions.
    %To the best of our knowledge, K-Radar is the first large-scale publicly-available 4DRT dataset and benchmark collected under diverse and challenging illumination, time, and weather conditions. 
    With the carefully annotated 3D bounding box labels and multimodal sensors, K-Radar can also be used for other autonomous driving tasks such as object tracking and odometry.
    \item We propose a 3D object detection baseline NN that directly consumes 4DRT as an input and verify that the height information of 4DRT is essential for 3D object detection.
    We also demonstrate the robustness of 4DRT-based perception for autonomous driving, especially under adverse weather conditions.
    \item We provide devkits that include: (1) training/evaluation, (2) labeling/calibration, and (3) visualization tools to accelerate 4DRT-based perception for autonomous driving research.
    %We provide the devkits that include: (1) training/evaluation code, (2) labeling/calibration tools, and (3) visualization tools to accelerate research in the field of 4DRT-based perception for autonomous driving.
\end{itemize}

The remaining of this paper is organized as follows. Section 2 introduces existing datasets and benchmarks that are related to perception for autonomous driving. Section 3 explains the K-Radar dataset and baseline NNs. Section 4 discusses the experimental results of the baseline NN on the K-Radar dataset. Section 5 concludes the paper with a summary and discussion on the limitations of this study.

\section{Related Works}

\begin{table}[h!]
\caption{Comparison of object detection datasets and benchmarks for autonomous driving. HR and LR refer to High Resolution Lidar with more than 64 channels and Low Resolution with less than 32 channels, respectively. Bbox., Tr.ID, and Odom. refer to bounding box annotation, tracking ID, and odometry, respectively. Bold text indicates the best entry in each category.}
%  \hl{Num. data, Bbox., Tr.ID, and Odom. refer to the number of labelled data, bounding box annotation, tracking ID, and odometry, respectively. Bolded text indicates the best entry in each category.}
\label{tab:comp_data_1}
\centering
\begin{tabular}{c|c|ccccc|ccc}
\hline\hline
Data                                                     & Num.  &    & \multicolumn{1}{r}{} & \multicolumn{1}{l}{Sensors} & \multicolumn{1}{r}{} & \multicolumn{1}{l|}{} &        & Label   &          \\
-set                                                     & data  & RT & RPC                  & LPC                        & Camera                       & GPS                   & Bbox. & Tr. ID & Odom. \\ \hline
\begin{tabular}[c]{@{}c@{}}K-Radar\\ (ours)\end{tabular} & 35K   & \textbf{4D} & \textbf{4D}                   & \textbf{HR.}                        & \textbf{360.}                         & \textbf{RTK}                   & \textbf{3D}     & \textbf{O}      & \textbf{O}        \\
VoD                                            & 8.7K  & X  & \textbf{4D}                   & \textbf{HR.}                        & Front                        & \textbf{RTK}                     & \textbf{3D}     & \textbf{O}      & \textbf{O}        \\         
Astyx                                                    & 0.5K  & X  & \textbf{4D}                   & LR.                        & Front                        & X                     & \textbf{3D}     & X      & X        \\
RADDet                                                   & 10K   & 3D & 3D                   & X                          & Front                        & X                     & 2D     & X      & X        \\
Zendar                                                   & 4.8K   & 3D & 3D                   & LR.                        & Front                        & GPS                   & 2D     & \textbf{O}      & O        \\
RADIATE                                                  & 44K   & 3D & 3D                   & LR.                        & Front                        & GPS                   & 2D     & \textbf{O}      & O        \\
CARRADA                                                  & 12.6K & 3D & 3D                   & X                          & Front                        & X                     & 2D     & \textbf{O}      & X        \\
CRUW                                                     & 396K  & 3D & 3D                   & X                          & Front                        & X                     & Point  & \textbf{O}      & X        \\
NuScenes                                                 & 40K   & X  & 3D                   & LR.                        & \textbf{360.}                         & \textbf{RTK}                   & \textbf{3D}     & \textbf{O}      & \textbf{O}        \\
Waymo                                                    & {230K}  & X  & X                    & \textbf{HR.}                        & \textbf{360.}                         & X                     & \textbf{3D}     & \textbf{O}      & X        \\
KITTI                                                    & 15K   & X  & X                    & \textbf{HR.}                        & Front                        & \textbf{RTK}                   & \textbf{3D}     & \textbf{O}      & \textbf{O}        \\
BDD100k                                                  & \textbf{120M}  & X  & X                    & X                          & Front                        & \textbf{RTK}                   & 2D     & \textbf{O}      & \textbf{O}        \\ \hline\hline
\end{tabular}
\end{table}

Deep neural networks generally require a large amount of training samples collected from diverse conditions so that they can achieve remarkable performance with excellent generalization. In autonomous driving, there are numerous object detection datasets that provide large-scale data of various sensor modalities, shown in Table \ref{tab:comp_data_1}.

\begin{wraptable}{O}{0.5\columnwidth}
\caption{Comparison of object detection datasets and benchmarks for autonomous driving. d/n refers to day and night. Bold text indicates the best entry in each category.}
\label{tab:conf_data_2}
\centering
\begin{tabular}{c|c|c}
\hline\hline
Dataset                                                  & Weather conditions                                                                  & Time \\ \hline
\begin{tabular}[c]{@{}c@{}}K-Radar\\ (ours)\end{tabular} & \textbf{\begin{tabular}[c]{@{}c@{}}overcast, fog,\\ rain, sleet, snow\end{tabular}} & \textbf{d/n}  \\
VoD                                                      & X                                                                                   & day \\
Astyx                                                    & X                                                                                   & day  \\
RADDet                                                   & X                                                                                   & day  \\
Zendar                                                   & X                                                                                   & day  \\
RADIATE                                                  & \begin{tabular}[c]{@{}c@{}}overcast, fog,\\ rain, snow\end{tabular}                 & \textbf{d/n}  \\
CARRADA                                                  & X                                                                                   & day  \\
CRUW                                                     & X                                                                                   & day  \\
NuScenes                                                 & overcast, rain                                                                      & \textbf{d/n}  \\
Waymo                                                    & overcast                                                                            & \textbf{d/n}  \\
KITTI                                                    & X                                                                                   & day  \\
BDD100k                                                  & \begin{tabular}[c]{@{}c@{}}overcast, fog,\\ rain, snow\end{tabular}                 & \textbf{d/n}  \\ \hline\hline
\end{tabular}
\end{wraptable}

KITTI \citep{kitti_metric} is one of the earliest and widely-used datasets for autonomous driving object detection that provide camera and Lidar measurements along with accurate calibration parameters and 3D bounding box labels. However, the number of samples and the diversity of the dataset is relatively limited since the 15K frames of the dataset are collected mostly in urban areas during daytime.
Waymo \citep{waymo} and NuScenes \citep{nuscenes} on the other hand provide a significantly larger number of samples with 230K and 40K frames, respectively. In both datasets, the frames are collected during both daytime and nighttime, increasing the diversity of the datasets.
Additionally, NuScenes provides 3D Radar point clouds (RPC), and \cite{rpc_fusion} demonstrates that utilizing Radar as an auxiliary input to the neural network can improve the detection performance of the network. However, RPC lose a substantial amount of information due to the CFAR thresholding operation and result in inferior detection performance when being used as the primary input to the network. For example, the state-of-the-art performance of Lidar-based 3D object detection on NuScenes dataset is 69.7\% mAP, whereas for Radar-based is only 4.9\% mAP.

In the literature, there are several 3DRT-based object detection datasets for autonomous driving. CARRADA \citep{carrada} provides Radar tensors in the range-azimuth and range-Doppler dimensions with labels of up to two objects in a controlled environment (wide flat surface). Zenar \citep{zendar}, RADIATE \citep{radiate}, and RADDet \citep{raddet} on the other hand provide Radar tensors collected on real road environments, but can only provide 2D BEV bounding box labels due to the lack of height information in 3DRTs. 
CRUW \citep{cruw} provides a large number of 3DRTs, but annotations only provide 2D point locations of objects.
VoD \citep{rpc_dataset} and Asytx \citep{astyx} provide 3D bounding box labels with 4DRPCs. However, the dense 4DRTs are not made available, and the number of samples in the datasets is relatively small (i.e., 8.7K and 0.5K frames). To the best of our knowledge, the proposed K-Radar is the first large-scale dataset that provide 4DRT measurements on diverse conditions along with 3D bounding box labels.

Autonomous cars should be capable to operate safely even under adverse weather conditions, therefore, the availability of adverse weather data in an autonomous driving dataset is crucial.
In the literature, the BDD100K \citep{bdd100k} and RADIATE datasets contain frames acquired under adverse weather conditions, as shown in Table \ref{tab:conf_data_2}.
%In the literature, the BDD100K \citep{bdd100k} and RADIATE datasets contain frames acquired under various adverse weather conditions.
However, BDD100K only provides RGB front images, while RADIATE only provides 32-channel low-resolution LPC.
%Unfortunately, the BDD100K is an RGB front image-only dataset, while RADIATE only provides 32-channel low-resolution LPC.
Meanwhile, the proposed K-Radar provides 4DRT, 64-channel and 128-channel high-resolution LPC, and 360-degree RGB stereo images, which enables the development of multi-modal approaches using Radar, Lidar, and camera for various perception problems for autonomous driving under adverse weather conditions.
%Meanwhile, the proposed K-Radar provides 4DRT, 64-channel and 128-channel high-resolution LPC, and 360-degree RGB stereo images
%Therefore, K-Radar enables the development of multi-modal approaches using Radar, Lidar, and camera for 3D object detection and other perception tasks for autonomous driving under adverse weather conditions.

\begin{figure}[b!]
% \begin{wrapfigure}{R}{0.7\columnwidth}
  \centering
  \includegraphics[width=1.0\columnwidth]{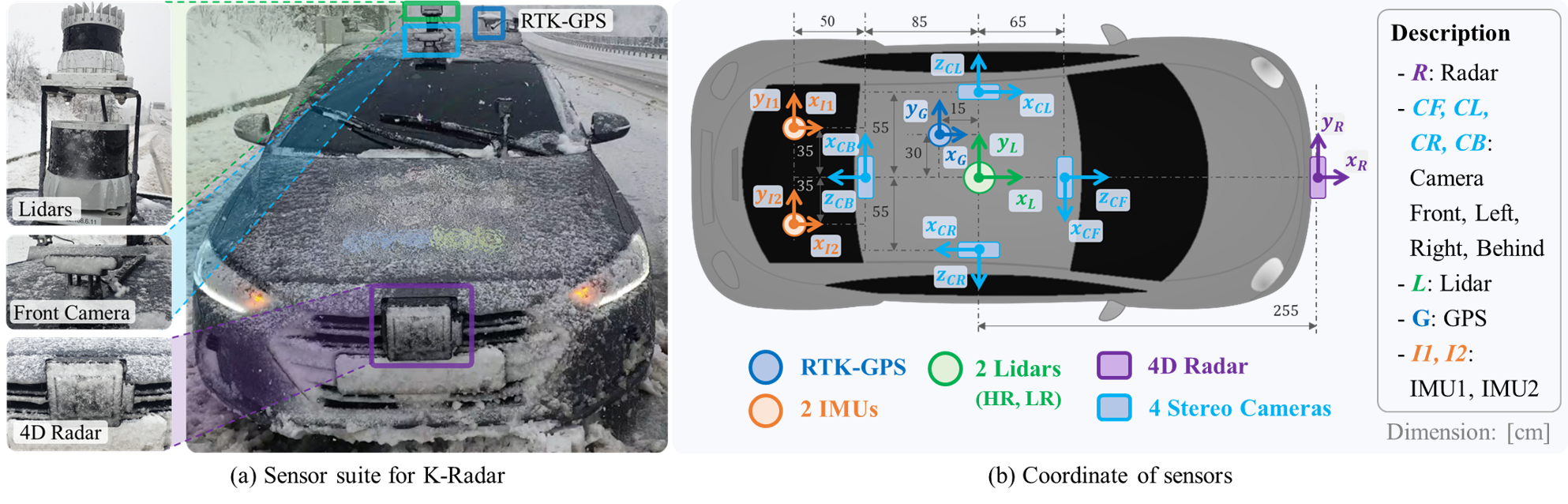}
  \caption{Sensor suite for K-Radar and coordinate system of each sensor. (a) shows the condition of the sensors after a 5 minute drive in heavy snow. Since the car drives forward, snow accumulates heavily in front of the sensors and covers the front camera lens, Lidar and Radar surfaces as shown in (a). As a result, during heavy snow, most of the information regarding the environment cannot be acquired by the front-facing camera and the Lidar. In contrast, Radar sensors are robust to adverse weathers, since the emitted waves can pass through raindrops and snowflakes. This figure emphasizes (1) the importance of Radar in adverse weather conditions, especially in heavy snowy conditions, and (2) the need for sensor placement and additional design (e.g., installation of wipers in front of the Lidar) considering the adverse weather conditions. (b) shows the installation location of each sensor and the coordinate system of each sensor.}
  \label{sensor_suite}
\end{figure}
% \end{wrapfigure}

\section{K-Radar}

%In this section, we introduce the K-Radar dataset and benchmark. 
In this section, we describe the configuration of the sensors used to construct the K-Radar dataset, the data collection process, and the distribution of the data. 
Then, we explain the data structure of a 4DRT, along with the visualization, calibration, and labelling processes. 
Finally, we present 3D object detection baseline networks that can directly consume 4DRT as the input.

\subsection{Sensor specification for K-Radar}

To collect data under adverse weathers, we install five types of waterproofed sensors ({listed in Appendix B}) with IP66 rating, according to the configuration shown in Figure \ref{sensor_suite}. First, a 4D Radar is attached to the front grill of the car to prevent multi-path phenomenon due to the bonnet or ceiling of the car. Second, a 64-channel Long Range Lidar and a 128-channel High Resolution Lidar are positioned at the centre of the car’s roof with different heights (Figure \ref{sensor_suite}-(a)). The Long-Range LPCs are used for accurately labelling objects of various distances, while the High-Resolution LPCs provide dense information with a wide (i.e., 44.5 degree) vertical field of view (FOV). Third, a stereo camera is placed on the front, rear, left, and right side of the vehicle, which results in four stereo RGB images that cover 360-degree FOV from the ego-vehicle perspective. Last, an RTK-GPS antenna and two IMU sensors are set on the rear side of the vehicle to enable accurate positioning of the ego-vehicle.

\subsection{Data collection and distribution}

\begin{figure}[h!]
  \centering
  \includegraphics[width=1.0\columnwidth]{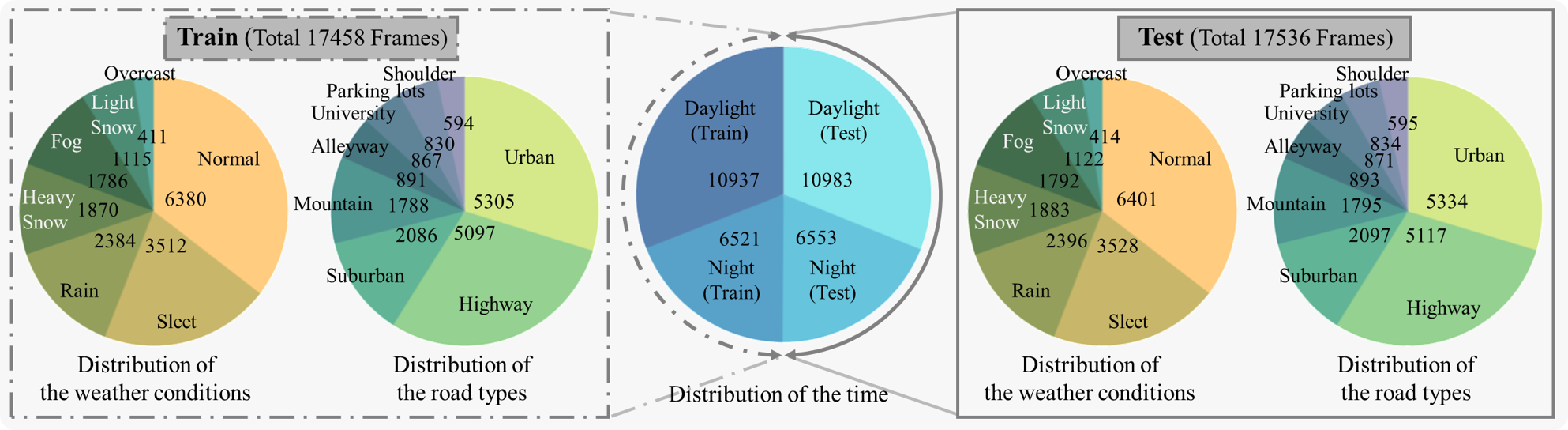}
  \caption{Distribution of data over collection time (night/day), weather conditions, and road types. The central pie chart shows the distribution of data over collection time, while the left and right pie charts show the distribution of data over weather conditions and road types for the train and test sets, respectively. At the outer edges of each pie chart, we state the collection time , weather conditions, and road types, and at the inner part, we state the number of frames in each distribution.}
  \label{dist_frame}
\end{figure}

\begin{wrapfigure}{R}{0.5\columnwidth}
%   \centering
  \includegraphics[width=0.5\columnwidth]{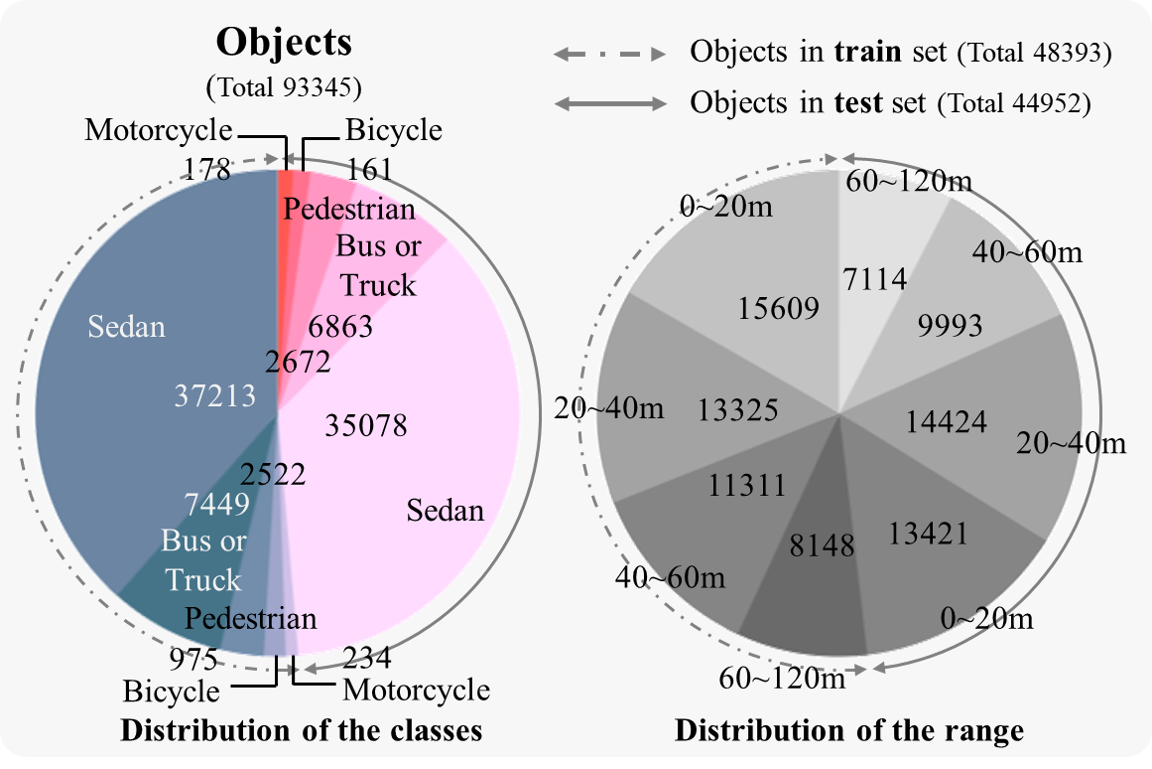}
  \caption{Objects classes and distance-to-ego-vehicle distribution for the training/test splits provided in the K-Radar dataset. We state the object class name and distance to the ego-vehicle in the outer part of the piechart, and the number of objects in each distribution in the inner part of the pie chart.}
  \label{dist_obj}
\end{wrapfigure}

The majority of frames with adverse weather conditions are collected in Gangwon-do of the Republic of Korea, a province that has the highest annual snowfall nationally. On the other hand, frames with urban environments are mostly collected in Daejeon of the Republic of Korea. The data collection process results in 35K frames of multi-modal sensor measurements that constitute the K-Radar dataset. We classify the collected data into several categories according to the criteria listed on Appendix C. In addition, we split the dataset into training and test sets in a way that each condition appears in both sets in a balanced manner, as shown in Figure \ref{dist_frame}.

In total, there are 93.3K 3D bounding box labels for objects (i.e., sedan, bus or truck, pedestrian, bicycle, and motorcycle) on the road within the longitudinal radius of 120m and lateral radius of 80m from the ego-vehicle. Note that we only annotate objects that appear in the positive longitudinal axis, i.e., in front of the ego-vehicle.

In Figure \ref{dist_obj}, we show the distribution of object classes and object distances from the ego-vehicle in the K-Radar dataset. The majority of objects lie within 60m distance from the ego-vehicle, with 10K$\sim$15K objects appearing in each of the 0m$\sim$20m, 20m$\sim$40m, and 40m$\sim$60m distance categories, and around 7K objects appearing in over 60m distance category. As a result, K-Radar can be used to evaluate the performance of a 3D object detection networks for objects on various distances.

\subsection{Data visualization, calibration, and annotation processes}

\begin{figure}[h]
  \centering
  \includegraphics[width=1.0\columnwidth]{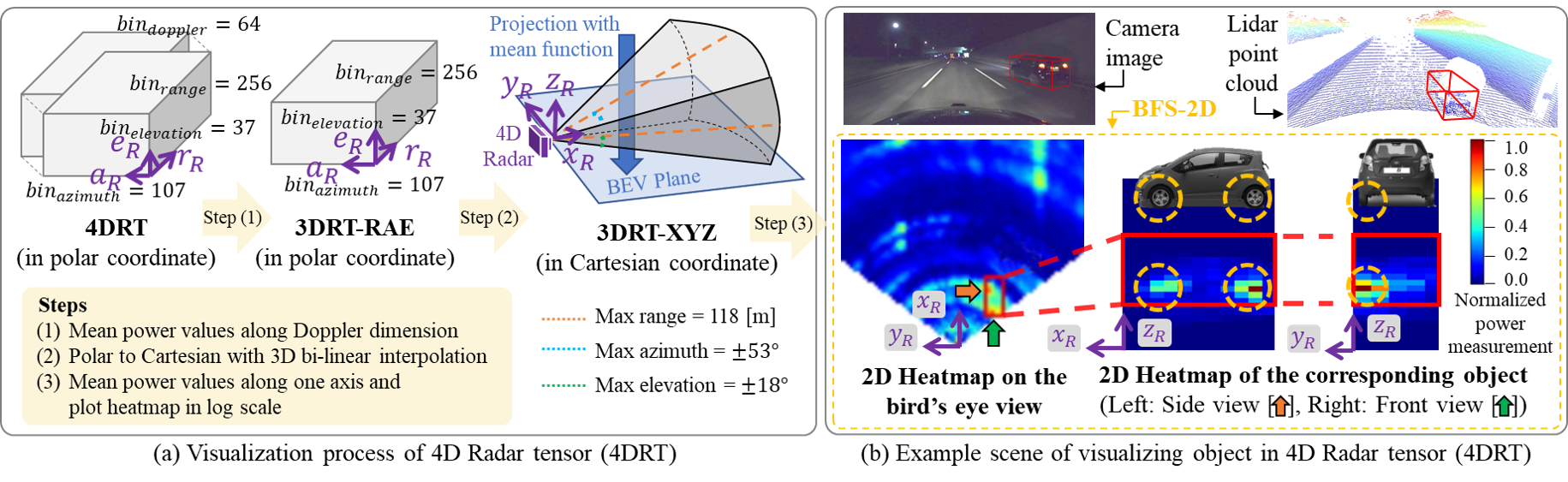}
  \caption{(a) A 4DRT visualization process and (b) the 4DRT visualization results. (a) is the process of visualizing 4DRT (polar coordinate) into BFS-2D (Cartesian coordinate) through a three-step process:
  (1) extracting the 3D Radar tensor that contains measurements along the range, azimuth, and elevation dimensions (3DRT-RAE) by reducing the Doppler dimension of the 4DRT through dimension-wise averaging,
  (2) transforming the 3DRT-RAE (polar coordinate) into 3DRT-XYZ (Cartesian coordinate),
%  \hl{of polar coordinate system into BFS-2D. The 4DRT-3D information is visualized as BFS-2D as shown in (b) through a three-step process shown in (a). The three steps of (a) are: (1) Extract the 3D Radar tensor that contains measurments along the range, azimuth, elevation dimensions (3DRT-RAE) from which the Doppler dimension of the 4DRT is removed. (2) Convert 3DRT-RAE to 3DRT (3DRT-XYZ) of the Cartesian coordinate system.} 
  (3) by removing one of the three dimensions of 3DRT-XYZ, the 4DRT is finally visualized as a two-dimensional Cartesian coordinate system. (b) is an example in which 4DRT-3D information is visualized as BFS-2D through the process of (a). We also show the front view camera image and the LPC of the same frame on the upper side of (b), and the bounding box of the car is marked in red. As shown in (b), the 4DRT is represented by three types of views (i.e., BEV, side view, and front view). We note that high power measurements are observed on wheels rather than the body of the vehicle when compared to the actual vehicle model picture with the side view and front view of the object. This is because radio wave reflection occurs mainly in wheels made of metal \citep{radar_wheel}, not in the body of a vehicle made of reinforced plastic.}
  \label{vis_process}
\end{figure}

Contrary to the 3D Radar tensor (3DRT) that lacks height information, 4D Radar tensor (4DRT) is a dense data tensor filled with power measurements in four dimensions: Doppler, range, azimuth, and elevation. However, the additional dimensionality of dense data imposes a challenge in visualizing 4DRT as a sparse data such as a point cloud (Figure \ref{kradar}). To cope with the problem, we visualize 4DRT as a two-dimensional heat map in the Cartesian coordinate system through heuristic processing as shown in Figure \ref{vis_process}-(a), which results in 2D heatmap visualizations in the bird-eye-view (BEV-2D), front-view (FV-2D), and side-view (SV-2D). We refer to these 2D heatmaps collectively as BFS-2D.

Through the BEV-2D, we can intuitively verify the robustness of 4D Radars to adverse weather conditions as shown in Figure \ref{kradar}. 
As mentioned earlier, camera and Lidar measurements can deteriorate under adverse weather conditions such as rain, sleet, and snow. 
In Figure \ref{kradar}-(e,f), we show that the measurements of a Lidar for a long-distance object are lost in heavy snow conditions. 
However, the BEV-2D of the 4DRT clearly indicate the object with high-power measurements on the edge of the bounding box of the objects.

Even with the BFS-2D, it is still challenging for a human annotator to recognize the shape of objects appearing on the frame and accurately annotate the corresponding 3D bounding boxes. 
Therefore, we create a tool that enables 3D bounding boxes annotation in LPCs where object shapes are more recognizable.
%Therefore, we create a tool that allows the annotators to annotate 3D bounding boxes on the Lidar point clouds where object shapes are more recognizable. 
In addition, we use the BEV-2D to help the annotators in the case of lost Lidar measurements due to adverse weather conditions. The details are covered in Appendix {D.1}.

We also present a tool for frame-by-frame calibration of the BEV-2D and the LPC to transform the 3D bounding box labels from the Lidar coordinate frame to the 4D Radar coordinate frame. The calibration tool supports a resolution of 1 cm per pixel with a maximum error of 0.5 cm. {The details of calibration between 4D Radar and Lidar are covered in Appendix D.2}.

{Additionally, we precisely obtain the calibration parameter between Lidar and the camera through a series of processes detailed in Appendix D.3. The calibration process between Lidar and camera enables the 3D bounding boxes and LPCs to be projected accurately onto camera images, which is crucial for multi-modal sensor fusion study, and can be used to produce dense depth maps for monocular depth estimation study.}

\subsection{Baseline NNs for K-Radar}

We provide two baseline NNs to demonstrate the importance of height information for 3D object detection: 
(1) Radar Tensor Network with Height (RTNH) that extracts feature maps (FMs) from RT with 3D Sparse CNN so that height information is utilized, and
%(1) Radar Tensor Network with Height (RTNH), which is the baseline NN extracting feature maps (FMs) from RT including height information with 3D Sparse CNN [ref] and 
(2) Radar Tensor Network without Height (RTN) that extracts FMs from RT with 2D CNN that does not utilize height information. 
%(2) Radar Tensor Network without Height (RTN), which is the other baseline NN extracting FMs from RT not including height information with 2D CNN. 
%This subsection describes the structure of RTNH and RTN, and the performance comparison results are covered in Section 4. 

\begin{figure}[h!]
  \centering
  \includegraphics[width=1.0\columnwidth]{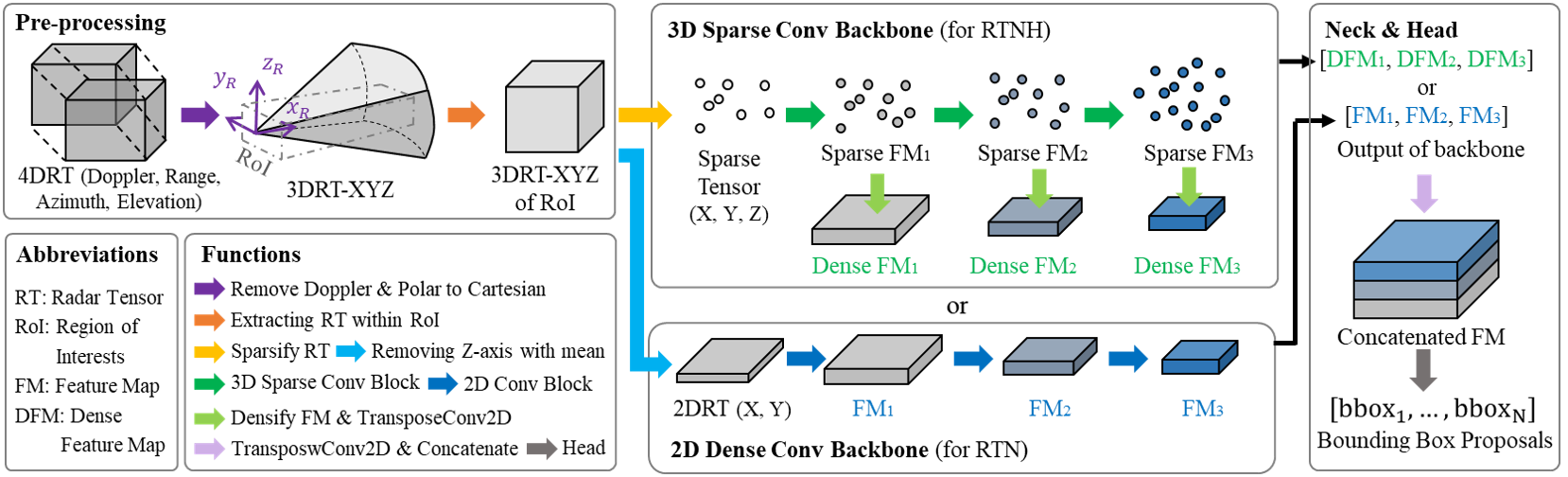}
  \caption{Two baseline NNs for verifying 4DRT-based 3D object detection performance.}
%  \caption{Two baseline NNs (with and without height information) for verifying 3D object detection performance.}
  \label{baseline_nns}
\end{figure}

As shown in Figure \ref{baseline_nns}, both RTNH and RTN consist of pre-processing, backbone, neck, and head. 
The pre-processing transforms the 4DRT from polar to Cartesian coordinate frame and extracts a 3DRT-XYZ within the region of interest (RoI). 
Note that we reduce the Doppler dimension by taking the mean value along the dimension. 
The backbone then extracts FMs that contain important features for the bounding box predictions.
%The backbone then extracts FMs to be used for the bounding box proposals.
And the head predicts 3D bounding boxes from the concatenated FM produced by the neck.
%The head extracts 3D bounding boxes from concatenated FM integrated by Neck. 

The network structure of RTNH and RTN, described in details in Appendix E, is similar except the backbone.
%RTNH and RTN are both composed of the same pre-processing, neck, head, and loss, and details of baseline NNs are described in Appendix C.
We construct the backbones of RTNH and RTN with 3D Sparse Conv Backbone (3D-SCB) and 2D Dense Conv Backbone (2D-DCB), respectively.
%We constructed the backbones of RTNH and RTN as 3D Sparse Conv Backbone (3D-SCB) and 2D Dense Conv Backbone (2D-DCB), respectively, in order to verify the performance of 3D object detection according to the presence of height information.
3D-SCB utilizes 3D sparse convolution \citep{sparse_conv} so that the three-dimensional spatial information (X, Y, Z) can be encoded into the final FM.
%3D-SCB of RTNH utilizes 3D convolution so that three-dimensional spatial information (X, Y, Z) including height information can be implied in the final FM. 
We opt to use the sparse convolution on sparse RT (top-10\% power measurements in the RT) since dense convolution on the original RT requires a prohibitively large amount of memory and computations that are unsuitable for real-time autonomous driving applications.
%However, since 3D convolution requires very large memory compared to 2D convolution, we constructed 3D-SCB with sparse convolution \citep{sparse_conv}, which utilizes only some elements (elements with top 10\% power measurements) of the input tensor. 
Unlike 3D-SCB, 2D-DCB uses 2D convolution so that only two-dimensional spatial information (X, Y) is encoded into the final FM.
%Unlike 3D-SCB, 2D-DCB uses 2D convolution so that only two-dimensional spatial information (X, Y) is implied to FM except for height information. 
As a result, the final FM produced by 3D-SCB contains 3D information (with height), whilst the final FM produced by 2D-DCB only contains 2D information (without height).
%In summary, 3D-SCB implies some elements of 3DRT-XYZ, including height information, to the final FM, and 2D-DCB implies all elements of 2DRT, except height information, to the final FM. 
%Details of the differences between 3D-SCB and 2D-DCB will be described in Appendix C.

\section{Experiment}

In this section, we demonstrate the robustness of 4DRT-based perception for autonomous driving under various weathers in order to find 3D object detection performance comparison between the baseline NN and a similarly-structured Lidar-based NN, PointPillars \citep{pointpillars}.
We also discuss the importance of height information by comparing 3D object detection performance between baseline NN with 3D-SCB backbone (RTNH) and baseline NN with 2D-DCB backbone (RTN).

%In this section, we use the K-Radar dataset to compare RTNH including height information and RTN not including it.
%Furthermore, we compare PointPillars \citep{pointpillars} with RTNH, one of the representative neural networks of Lidar 3D object detection, on the K-Radar dataset.
%The comparison results highlight the significance of 4DRT height information in 3D object detection and the robustness of 4D Radar over Lidar in adverse weather conditions.

\subsection{Experiment Setup and Metric}

%\begin{wraptable}{R}{0.42\linewidth}
%    \centering
%    \label{tab:hyperparams}
%    \caption{Hyper-parameters for training}
%    \begin{tabular}{cc}
%    \hline
%    Parameters    & Value             \\ \hline\hline
%    Batch size    & 24                \\ \hline
%    Learning rate & 1e-3              \\ \hline
%    Optimizer     & Adam              \\ \hline
%    Scheduler     & CosineAnnealingLR \\ \hline
%    Max epoch     & 10                \\ \hline
%    \end{tabular}
%\end{wraptable}

\textbf{Implementation Detail}
We implement the baseline NNs and PointPillars using PyTorch 1.11.0 on Ubuntu machines with a RTX3090 GPU.
We set the batch size to 4 and train the networks for 11 epochs using Adam optimizer with a learning rate of 0.001.
Note that we set the detection target to the sedan class, which has the largest number of samples in K-Radar dataset.

\begin{wraptable}[8]{R}{0.48\columnwidth}
\caption{Performance comparison of baseline NNs with or without height Information.}
\label{tab:rtnh}
\centering
\begin{tabular}{c|ccc}
\hline\hline
\begin{tabular}[c]{@{}c@{}}Baseline\\ NNs\end{tabular} & \begin{tabular}[c]{@{}c@{}}$AP_{3D}$\\ $[\%]$\end{tabular} & \begin{tabular}[c]{@{}c@{}}$AP_{BEV}$\\ $[\%]$\end{tabular} & \begin{tabular}[c]{@{}c@{}}GPU RAM\\ $[MB]$\end{tabular} \\ \hline
RTNH                                                   & \textbf{47.44}                                          & \textbf{58.39}                                          & \textbf{421}                                              \\
RTN                                                    & 40.12                                          & 50.67                                          & 520                                              \\ \hline\hline
\end{tabular}
\end{wraptable}
%\textbf{Implementation Details} We train RTNH, RTN, and Point Pillars using the RTX3090 GPU on the K-Radar dataset for 10 epochs. 
%We applied the hyper-parameters for training equally to the neural network as shown in Table 4, and all training and evaluation are performed in Pytorch1.11.0 of Ubuntu 18.04 machine. 
%The target object is set to a sedan, which occupies the largest distribution in K-Radar dataset.
%For all experiments, we set the batch size as 24 and use Adam optimizer with 0.001 initial learning rate, cosine annealing scheduler, and a maximum epoch of 10.

\textbf{Metric}
In the experiments, we utilize the widely-used Intersection Over Union (IOU)-based Average Precision (AP) metric to evaluate the 3D object detection performance.
We provide APs for BEV ($AP_{BEV}$) and 3D ($AP_{3D}$) bounding boxes predictions as in \citep{kitti_metric}, where a prediction is considered to be a true positive if the IoU is over 0.3.

\subsection{Comparison between RTN and RTNH}

%We evaluate the performance of the neural networks with Intersection of Union (IoU)-based Average Precision, a commonly used performance evaluation metric in 3D object detection. 
%We evaluate object detection performance by comparing APs for BEVs and 3D bounding boxes ($AP_{BEV}$ and $AP_{3D}$, respectively), as in \citep{kitti_metric} with 0.3 IoU threshold. 
%We unified the coordinate system and RoI of all the neural networks for fair evaluation.

% \begin{wraptable}[15]{R}{0.3\columnwidth}
% \caption{Performance comparison of baseline NNs with or without height Information.}
% \label{tab:rtnh}
% \begin{tabular}{c|cc}
% \hline\hline
% \begin{tabular}[c]{@{}c@{}}Baseline\\ NNs\end{tabular}      & RTNH           & RTN   \\ \hline
% \begin{tabular}[c]{@{}c@{}}$AP_{3D}$\\ $[\%]$\end{tabular}  & \textbf{41.05} & 31.62 \\
% \begin{tabular}[c]{@{}c@{}}$AP_{BEV}$\\ $[\%]$\end{tabular} & \textbf{54.92} & 52.96 \\
% \begin{tabular}[c]{@{}c@{}}GPU RAM\\ $[MB]$\end{tabular}    & \textbf{371}   & 506   \\ \hline\hline
% \end{tabular}
% \end{wraptable}

We show the detection performance comparison between RTNH and RTN on Table \ref{tab:rtnh}.
%We compare the performance of RTNH and RTN trained with all frames of K-Radar including adverse weather conditions. 
We can observe that RTNH has 7.32\% and 7.72\% higher performance in $AP_{3D}$ and $AP_{BEV}$, respectively, compared to RTN.
%As can be seen from Table 4, RTNH has 9.43\% and 1.96\% higher performance in $AP_{3D}$ and $AP_{BEV}$, respectively, than RTN. 
RTNH significantly surpasses RTN in terms of both $AP_{3D}$ and $AP_{BEV}$, indicating the importance of height information available in the 4DRT for 3D object detection.
%In particular, it can be seen that RTNH has significantly improved performance at $AP_{3D}$ compared to RTN, and height information of 4DRT is important in 3D object detection. 
Furthermore, RTNH requires less GPU memory compared to RTN since it utilizes the memory-efficient sparse convolutions as mentioned in Section 3.4.
%Furthermore, the results in Table 4 show that RTNH performs better than RTN while using less GPU RAM, as RTNH applies sparse convolution that utilizes only some elements of 4DRT as mentioned in Section 3.4.

\subsection{Comparison between RTNH and PointPillars}

\begin{table}[htb!]
\centering
\caption{Performance comparison of NNs of Radar and Lidar under various weather conditions}
\label{tab:lidar_radar}
\begin{tabular}{c|c|llllllll}
\hline\hline
Networks & Metric & Total         & \begin{tabular}[c]{@{}l@{}}nor-\\ mal\end{tabular} & \begin{tabular}[c]{@{}l@{}}over-\\ cast\end{tabular} & fog           & rain          & sleet         & \begin{tabular}[c]{@{}l@{}}light\\ snow\end{tabular} & \begin{tabular}[c]{@{}l@{}}heavy\\ snow\end{tabular} \\ \hline
RTNH    & $AP_{3D} [\%]$     & \textbf{47.4} & 49.9    & \textbf{56.7} & \textbf{52.8} & 42.0 & \textbf{41.5} & \textbf{50.6} & \textbf{44.5} \\
(4D Radar) & $AP_{BEV} [\%]$ & \textbf{58.4} & \textbf{58.5}    & \textbf{64.2} & \textbf{76.2} & \textbf{58.4} & \textbf{60.3} & \textbf{57.6} & \textbf{56.6} \\ \hline
PointPillars & $AP_{3D} [\%]$ & 45.4 & \textbf{52.3} & 56.0 & 42.2 & \textbf{44.5} & 22.7 & 40.6 & 29.7 \\
(Lidar)      & $AP_{BEV} [\%]$ & 49.3 & 56.6 & 61.0 & 52.0 & 57.8 & 23.1 & 51.6 & 30.8 \\ \hline\hline
\end{tabular}
\end{table}

We show the detection performance comparison between RTNH and a similarly-structured Lidar-based detection network, PointPillars, in Table \ref{tab:lidar_radar}. 
%We compare the performance of RTNH and PointPillars, one of the representative neural networks of Lidar 3D object detection, under various weather conditions including adverse weathers. 
%As shown in Table 5, the performance results $AP_BEV$ and $AP_3D$ of RTNH are 21.7\% and 18.3\% higher than PoinPillars in total, respectively. 
The Lidar-based network suffers significant BEV and 3D detection performance drops of 33.5\% and 29.6\% or 25.8\% and 22.6\%, respectively, in sleet or heavy snow condition compared to the normal condition.
In contrast, the 4D radar-based RTNH detection performance is hardly affected by adverse weathers, where the BEV and 3D object detection performances in sleet or heavy snow condition are better or similar compared to the normal condition.
The results testify the robustness of 4D radar-based perception in adverse weathers.
%In particular, when comparing $AP_BEV$ and $AP_3D$ from total to heavy snow, PointPillars decreases by 11.4\% and 17.9\%, respectively, but RTNH shows similar or high performance. 
%From the results of Table 5, PointPillars performs poorly in adverse weather, especially heavy snow, whereas RTNH shows almost the same performance in heavy snow. 
%In this experiment, we do not show that our baseline NN is a better neural network compared to PointPillars, but demonstrate that 4D Radar is more robust than Lidar under adverse weathers, and the results in Table 6 prove this.
We provide qualitative results and additional discussions for other weather conditions in Appendix F.
%Appendix F has additional discussions for other weather conditions including qualitative results.

\section{Limitation and Conclusion}
In this section, we discuss the limitations of K-Radar and provide a summary of this work, along with suggestions on the future research directions.
%In this section, we discuss limitations of K-Radar and discuss the final conclusions of this paper.

\subsection{Limitation of the FOV coverage of 4DRTs}
As mentioned in Section 3.1, K-Radar provides 4D radar measurements in the forward direction, with an FOV of 107 degree.
The measurement coverage is more limited compared to the 360 degree FOV of Lidar and camera.
This limitation is originated from the size of a 4DRT with dense measurements in four dimensions, which require significantly larger memory to store the data compared to a camera image with two dimensions or a LPC with three dimensions.
Specifically, the size of the 4DRT data in K-Radar is roughly 12TB, while the size of surround camera images data is about 0.4TB, and the size of LPCs data is about 0.6TB.
Since providing 360 degrees 4DRT measurements requires a prohibitively large amount of memory, we opt to record 4DRT data only in the forward direction, which could provide the most relevant information for autonomous driving.
%K-Radar provides LPC and camera images measuring 360 degrees around the ego-vehicle, but 4D Radar provides only measurements obtained from a 107 degree field of view (FOV) installed on the front grill, as mentioned in Section 3.1. 
%The reason for this is as follows: 4DRT as a total of 64 x 256 x 107 x 37 (= 65M) power measurements per frame, with a capacity of approximately 300 MB. 
%Camera images are 342.2GB, LPC is 638GB, 4DRT is 10.7TB, and 4DRT is 91.6 percent of the total data capacity in a total of 35K frames. 
%As a result, providing 4DRT measuring 360 degrees roughly requires 10.7 x 4 = 42.8TB. 
%As such, we mainly provide the forward information, which is the most important information for driving planning.

%\subsection{Baseline NN utilizing only height information}
%Because Doppler measurement represents the relative velocity of other objects relative to the ego-vehicle, it may be useful for detecting and tracking moving objects in 3D object detection neural networks. 
%However, 4DRT adds elevation information to 3DRT, which in this study focuses on the need for height information that 3DRT lacks, and the baseline NN we present concentrates on 3D object detection utilizing height information. 
%We note that the K-Radar dataset includes all of Doppler measurement, and we expect to refer this in the future when utilizing the K-Radar dataset.

\subsection{Conclusion}
In this paper, we have introduced a 4DRT-based 3D object detection dataset and benchmark, K-Radar. 
The K-Radar dataset consists of 35K frames with 4DRT, LPC, surround camera images, and RTK-IMU data, all of which are collected in various time and weather conditions. 
%The K-Radar dataset is collected in a variety of weather conditions, including adverse weathers during the day and night, and consists of a total of 35K frames of 4DRT and calibrated LPC, surrounded camera images, and RTK-GPS/IMU data. 
K-Radar provides 3D bounding box labels and tracking ID for {93.3K} objects of five classes with distance of up to 120 m.
%In the K-Radar dataset, a total of 93300 objects divided into five objects are labeled as 3D bounding boxes, and the ego-vehicle and object distances are evenly distributed between 0 and 120m. 
To verify the robustness of 4D radar-based object detection, we introduce baseline NNs that uses 4DRT as the input.
%In addition, we introduce baseline NN based on 4DRT and validate its performance using the K-Radar dataset. 
From experimental results, we demonstrate the importance of height information that is not available in the conventional 3DRT and the robustness of 4D radar under adverse weathers.
%Comparing the performance of RTNH using height information of 4DRT and RTN not using it, we confirm the importance of 4DRT height information in 3D object detection. 
%Furthermore, we compare the performance of RTNH and PointPillars, a representative neural network of Lidar 3D object detection, demonstrating that the Radar sensor is a robust sensor in adverse weather. 
While the experiments in this work are focused on 4DRT-based 3D object detection, K-Radar can be used for 4DRT-based object tracking, SLAM, and various other perception tasks.
Therefore, we hope that K-Radar can accelerate works in 4DRT-based perception for autonomous driving.
%The baseline NN provided in this paper focuses on 3D object detection using height information of 4DRT, and it may be considered to provide baseline NN for 4DRT-based object tracking and SLAM, etc. in the future. 
%We hope that K-Radar provided in this paper will be widely used in neural network research using 4DRT, and that 4DRT-based research will help to proceed actively.

\section*{Acknowledgment}

This work was partly supported by Institute of Information \& communications Technology Planning \& Evaluation (IITP) grant funded by the Korea government (MSIT) (No. 01210790) and the National Research Foundation of Korea (NRF) grant funded by the Korea government (MSIT) (No. 2021R1A2C3008370). \\

\appendix

\newcounter{appdxTableint}
\newcounter{appdxFigureint}
\newcommand\tabcounterint{%
  \refstepcounter{appdxTableint}%
  \renewcommand{\thetable}{\arabic{appdxTableint}}%
}
\newcommand\figcounterint{%
  \refstepcounter{appdxFigureint}%
  \renewcommand{\thefigure}{\arabic{appdxFigureint}}%
}
\setcounter{appdxFigureint}{7}
\setcounter{appdxTableint}{4}

\textbf{Appendix}
The appendix is organized as follows.
We present details of the K-Radar dataset, the sensor suite, and criteria of conditions (weather {conditions}, road structures, and collection time) in Section A, B, and C, respectively. 
We also provide details of the annotation/calibration process and the baseline neural networks (NNs) in Section D and E, respectively. 
{We discuss results regarding each weather condition and consideration of the K-Radar dataset as a pre-training dataset for other Radar tensor datasets in Section F and G, respectively.}
In addition, we report performance of RTNH on wider areas and multiple classes in Section H.
Finally, we introduce details of devkits and list relevant URLs to help with understanding the content of the paper in Section I and J, respectively.

\section{Additional details for K-Radar dataset}

In this section, we present additional samples of K-Radar dataset, sequence distribution, dataset composition, license, and privacy concerns.

\begin{figure}[b!]
{
  \figcounterint
  \centering
  \includegraphics[width=1.0\columnwidth]{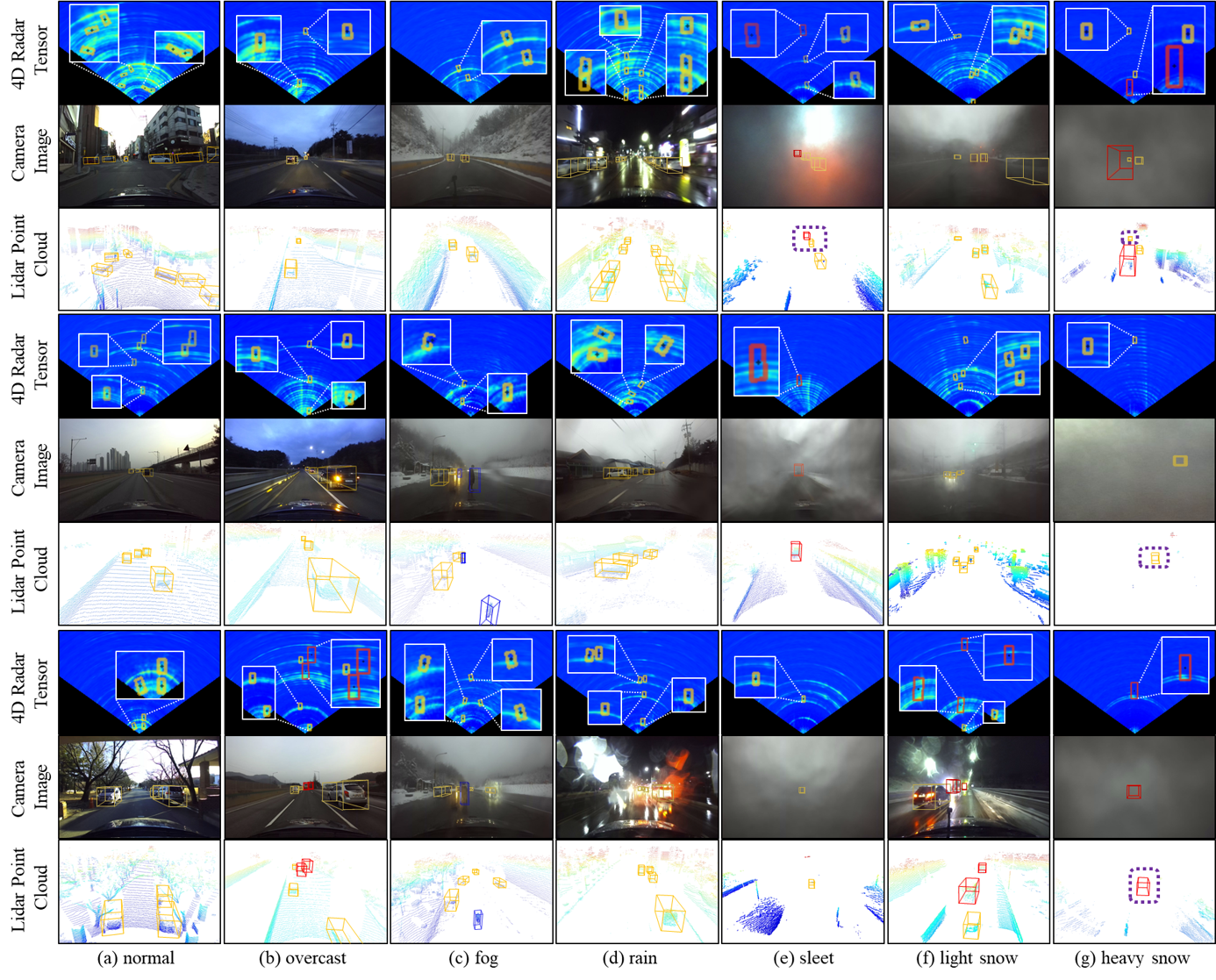}
  \caption{Additional samples of K-Radar datasets for various weather conditions. (1) 4DRTs, (2) front view camera images, and (3) Lidar point clouds (LPCs) of three different road conditions with the same weather condition are depicted in three boxes in each column. In this example, yellow, red, and blue bounding boxes represent the sedan, bus or truck, and pedestrian classes, respectively. Objects with all LPC measurements missing due to the adverse weather are marked with purple dotted lines. 
  More samples of K-Radar dataset can be visualized using the devkits program described in Section I.}
  %In addition to the samples shown in Figure 2 and Figure 8, samples of different conditions can be found using the K-Radar dataset and the devkits program described in Section I.}
  \label{app_kradar}
}
\end{figure}

\begin{wrapfigure}{R}{0.62\columnwidth}
{
  \figcounterint
  \centering
  \includegraphics[width=0.62\columnwidth]{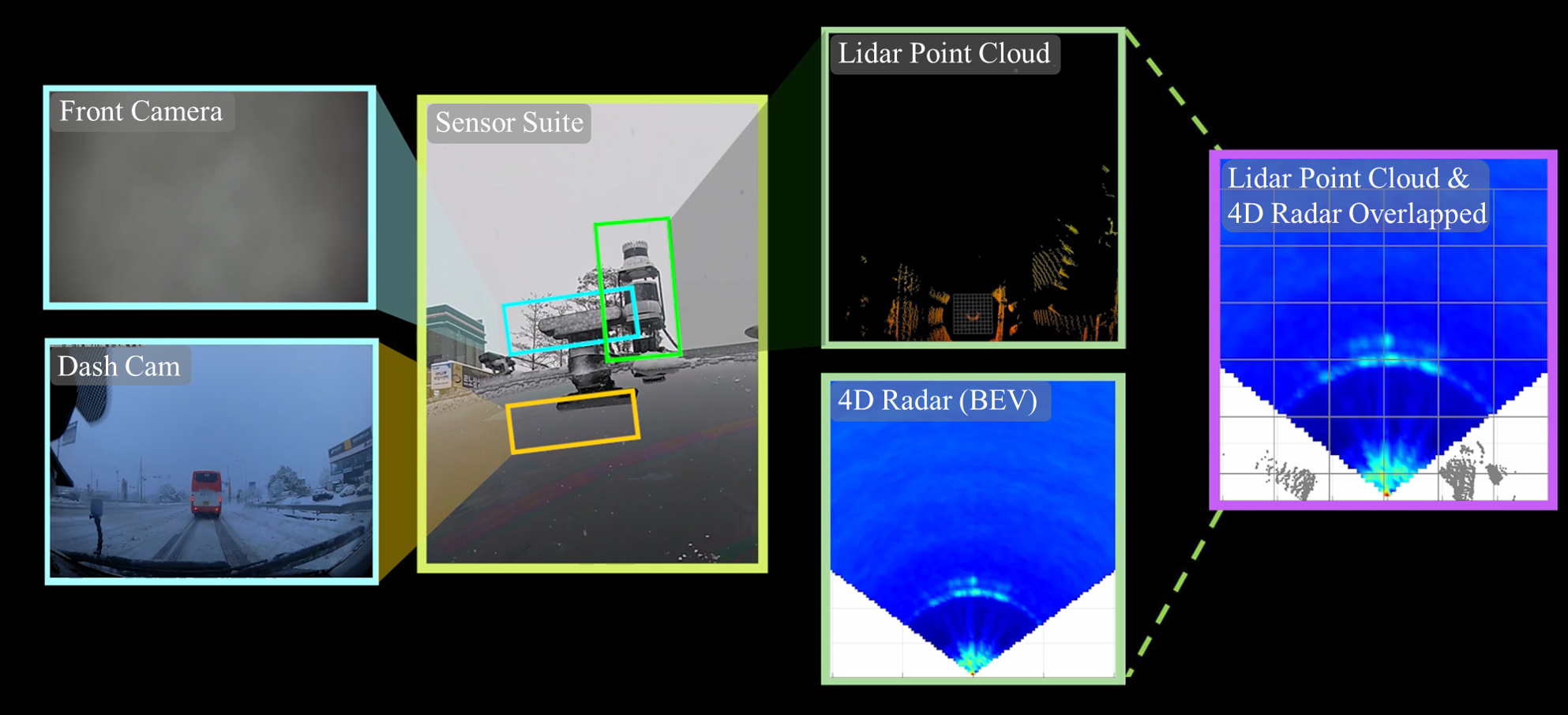} % 0.62
  \caption{{A snippet} of the video clip that shows each sensor measurement dynamically changing during driving under the heavy snow condition. (see Section J URL 2)}
  \label{heavy_snow}
}
\end{wrapfigure}

\subsection{Additional samples of the K-Radar dataset and explanation of LPCs for each weather condition}

In the sleet (Figure \ref{app_kradar}-(e)) or heavy snow (Figure \ref{app_kradar}-(g)) condition, the Lidar point cloud (LPC) measurements of some objects ahead are lost when the ego-vehicle is driving. 
Conversely, in the rain (Figure \ref{app_kradar}-(d)) or light snow (Figure \ref{app_kradar}-(f)) condition, LPC measurements of objects exist.
%Conversely, in the light snow (Figure \ref{app_kradar} condition, LPC measurements of objects exist at all.
%%% Are we sure the reason is this? Or just one of the possible explanations? %%%%%%%%
The reason for this is as follows.
Sleet is a mix of rain and snow that freezes when it falls from the sky in a liquid state and comes into contact with a sensor or an ego-vehicle colder than the air \citep{sleet}.
%Sleet is a mix of rain and snow, freezing when it falls from the sky in a liquid state and comes into contact with a sensor or an ego-vehicle colder than the air \citep{sleet}. 
% I don't understand the next sentence meaning%
In our case, the sleet freezes on the front surface of the Lidar sensor, creating a layer of frost.
%In particular, most of the front surface of the ego-vehicle driving forward touches the sleet, and in the case of Lidar sensor, most of the front surface freezes.
Thus, as shown in Figure \ref{app_kradar}-(e), the measuring signals from the Lidar cannot reach objects in the front of the ego-vehicle, which results in missing points in the LPC.
%Thus, as shown in Figure 8-(e), some LPC measurements of objects in front of the ego-vehicle are not measured.
In addition, heavy snow is a weather condition in which snow falls over 1 cm per hour as described in Table \ref{tab:app_criteria}, and Figure 3 shows that a lot of snow accumulates on the front surface of the sensor after the vehicle drives forward for 5 minutes.
For this reason, similar to sleet, some LPC measurements of objects in front of the ego-vehicle are missing, as shown in Figure \ref{app_kradar}-(g).
%For this reason, similar to sleet, some LPC measurements of objects in front of the ego-vehicle are not measured, as shown in Figure \ref{app_kradar}-(g).
Unlike sleet and heavy snow, there is only a little-to-no amount of snow accumulation on the front surface of the Lidar sensor in light snow condition, as described in Table 8.
In addition, LPC measurements of objects are also partially available in rain condition, since raindrops slip over the front surface of the Lidar sensor.
%Unlike sleet and heavy snow, light snow is not so much that snow accumulates on the front surface of the Lidar sensor, as described in Table 8.
Therefore, the Lidar sensors can measure objects in front of the ego-vehicle as shown in Figure \ref{app_kradar}-(d) and (f).
%As such, it can be seen that all LPC measurements of objects in front of the ego-vehicle are measured as shown in Figure \ref{app_kradar}-(f).
Note that we provide a video clip (Figure \ref{heavy_snow}) in Section J URL 2 to show sensor measurements collected while driving forward under the heavy snow condition.
%As shown in Figure \ref{heavy_snow}, we provide a movie to show sensor measurements collected while driving forward under the heavy snow condition, and it is provided in Section K URL 2.

% \begin{figure}[t!]
% {
%   \figcounterint
%   \centering
%   \includegraphics[width=0.6\columnwidth]{fig9.png}
%   \caption{{A snippet} of the video clip that shows each sensor measurement dynamically changing during driving under the heavy snow condition. (see Section K URL 2)}
%   \label{heavy_snow}
% }
% \end{figure}

% To summarize, the LPC measurements of objects in front of the ego-vehicle are lost in sleet and heavy snow conditions since the surface of the Lidar sensor is covered by frost or accumulated snow.
%To summarize, the LPC measurements of the objects, which is located in the part where the surface of Lidar sensor is covered, are lost in sleet or heavy snow condition, since the surface of the Lidar sensor freezes or a lot of snow accumulates.
% On the other hand, in the light snow condition, measurements of all objects exist, since the Lidar sensor surface is not covered by frost or accumulated snow.
%On the other hand, in light snow, measurements of all objects exist because the Lidar sensor surface does not freeze or accumulate snow, and the sensor surface is not covered, and this can be seen in most of the K-Radar dataset.

%%%%%%%%%%%%%%%%%%%%%%%%% Until Here %%%%%%%%%%%%%%%%%%%%%%%%%%%%%%%%%%%%%%%%%
\subsection{Sequence distribution}

The K-Radar dataset provides a total of 35K frame data obtained in different weather conditions, road structures, and collection time.
The dataset is divided into 58 sequences, where the details of each sequence are shown in Table \ref{tab:app_seq_dist}.
%summarizes the details of each sequence (number of frames, weather conditions, road structures, collection time). 
%For dataset deployment, we compress the 58 sequences into 12 compressed files, with further explanation in Section A.3.
%For the convenience of K-Radar dataset deployment, we provide a dataset by compressing 58 sequences, and the details for each compressed file are provided in detail in Section A.3.

\subsection{Dataset composition}

Each of the 58 sequences consists of 12 compressed folders, as shown in Table \ref{tab:app_composition}. 
Table \ref{tab:app_composition} provides information on the folder name, data type, extension, size, and the usage of each folder.

\subsection{License}

The K-Radar dataset is published under the CC BY-NC-ND License, and all codes are {published} under the Apache License 2.0.

% \subsection{Additional information for the related datasets}

% As shown in Table \ref{tab:app_add_info}, {we provide the additional information of the related datasets which is not listed in Section 2, and possibly useful.}

\subsection{Privacy concerns}

\begin{figure}[h!]
{
  \figcounterint
  \centering
  \includegraphics[width=1.0\columnwidth]{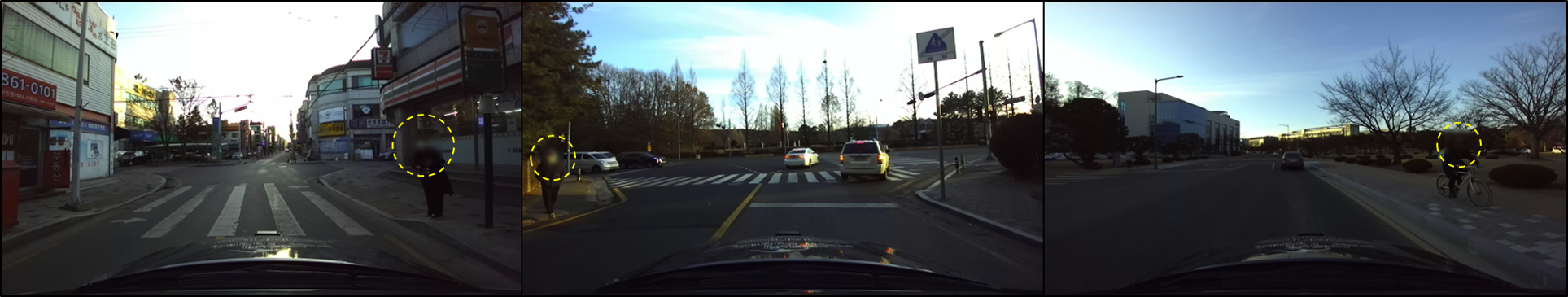}
  \caption{{Examples of front images showing people whose faces are blurred.}}
  \label{privacy_concerns}
}
\end{figure}

We confirm that all the image sequences with pedestrians, bicycles, and motorcycles do not have recognizable faces. Although everyone is wearing a mask due to COVID-19, we have taken additional precautions and blurred all faces to protect their privacy as shown in Figure \ref{privacy_concerns}.

\begin{table}[t!]
{
\tabcounterint
\begin{center}
\caption{Sequence of the K-Radar dataset; sequences 1 through 20 are obtained {in} Dae-jeon, and sequences 21 through 58 are obtained {in} Gang-won Province. `he. snow' and `park.lot' denotes heavy snow and parking lot, respectively.}
\label{tab:app_seq_dist}
\begin{tabular}{ccccc||ccccc} \hline\hline
Seq. & \begin{tabular}[c]{@{}c@{}}Num.\\ Fr.\end{tabular} & \begin{tabular}[c]{@{}c@{}}Weather\\ Cond.\end{tabular} & \begin{tabular}[c]{@{}c@{}}Road\\ Stru.\end{tabular} & Time  & Seq. & \begin{tabular}[c]{@{}c@{}}Num.\\ Fr.\end{tabular} & \begin{tabular}[c]{@{}c@{}}Weather\\ Cond.\end{tabular} & \begin{tabular}[c]{@{}c@{}}Road\\ Stru.\end{tabular} & Time  \\ \hline
1    & 597                                                & normal                                                  & urban                                                & night & 30   & 470                                                & sleet                                                   & park.lot                                             & day   \\
2    & 462                                                & normal                                                  & highway                                              & night & 31   & 598                                                & sleet                                                   & suburban                                             & day   \\
3    & 597                                                & normal                                                  & highway                                              & night & 32   & 597                                                & rain                                                    & suburban                                             & day   \\
4    & 588                                                & normal                                                  & highway                                              & night & 33   & 598                                                & rain                                                    & suburban                                             & day   \\
5    & 597                                                & normal                                                  & urban                                                & day   & 34   & 598                                                & rain                                                    & suburban                                             & night \\
6    & 594                                                & normal                                                  & urban                                                & night & 35   & 597                                                & sleet                                                   & park.lot                                             & night \\
7    & 595                                                & normal                                                  & alleyway                                             & night & 36   & 597                                                & sleet                                                   & park.lot                                             & night \\
8    & 567                                                & normal                                                  & university                                           & night & 37   & 597                                                & sleet                                                   & suburban                                             & night \\
9    & 833                                                & normal                                                  & highway                                              & day   & 38   & 597                                                & fog                                                     & mountain                                             & day   \\
10   & 1130                                               & normal                                                  & highway                                              & day   & 39   & 597                                                & fog                                                     & mountain                                             & day   \\
11   & 1195                                               & normal                                                  & highway                                              & day   & 40   & 598                                                & fog                                                     & mountain                                             & day   \\
12   & 888                                                & normal                                                  & highway                                              & day   & 41   & 597                                                & fog                                                     & mountain                                             & day   \\
13   & 227                                                & overcast                                                & highway                                              & day   & 42   & 598                                                & light snow                                                & urban                                                & day   \\
14   & 595                                                & normal                                                  & urban                                                & day   & 43   & 598                                                & light snow                                                & urban                                                & day   \\
15   & 591                                                & normal                                                  & urban                                                & day   & 44   & 597                                                & fog                                                     & shoulder                                             & day   \\
16   & 578                                                & normal                                                  & university                                           & day   & 45   & 592                                                & fog                                                     & shoulder                                             & day   \\
17   & 593                                                & normal                                                  & university                                           & day   & 46   & 598                                                & he. snow                                                & highway                                              & night \\
18   & 594                                                & normal                                                  & urban                                                & day   & 47   & 266                                                & he. snow                                                & highway                                              & night \\
19   & 592                                                & normal                                                  & alleyway                                             & day   & 48   & 443                                                & light snow                                                & highway                                              & night \\
20   & 595                                                & normal                                                  & urban                                                & day   & 49   & 598                                                & light snow                                              & highway                                              & night \\
21   & 597                                                & rain                                                    & alleyway                                             & night & 50   & 597                                                & sleet                                                   & highway                                              & night \\
22   & 598                                                & overcast                                                & urban                                                & night & 51   & 597                                                & sleet                                                   & highway                                              & night \\
23   & 598                                                & rain                                                    & urban                                                & night & 52   & 598                                                & sleet                                                   & highway                                              & night \\
24   & 598                                                & rain                                                    & urban                                                & night & 53   & 597                                                & sleet                                                   & highway                                              & day   \\
25   & 597                                                & rain                                                    & urban                                                & night & 54   & 601                                                & he. snow                                                & urban                                                & day   \\
26   & 597                                                & rain                                                    & suburban                                             & day   & 55   & 494                                                & he. snow                                                & urban                                                & day   \\
27   & 598                                                & sleet                                                   & suburban                                             & day   & 56   & 598                                                & he. snow                                                & urban                                                & day   \\
28   & 597                                                & sleet                                                   & mountain                                             & day   & 57   & 598                                                & he. snow                                                & urban                                                & day   \\
29   & 597                                                & sleet                                                   & mountain                                             & day   & 58   & 598                                                & he. snow                                                & urban                                                & day  \\\hline\hline
\end{tabular}
\end{center}
}
\end{table}

\begin{table}[t!]
{
\tabcounterint
\begin{center}
\caption{Dataset composition of each sequence. `res.', `cam.', and `img.' denotes resolution, camera, and image, respectively.}
\label{tab:app_composition}
\begin{tabular}{ccccc} \hline\hline
folder name     & data type          & extension & size & usage                                  \\ \hline
radar\_tesseract & 4DRT               & .mat      & 360GB         & network input, visualization \\
radar\_xyz\_cube  & 3DRT-XYZ           & .mat      & 72GB          & network input                          \\
os1-128         & High res. LPC      & .pcd      & 14GB          & network input, visualization           \\
os2-64          & Low res. LPC       & .pcd      & 7GB           & network input, visualization           \\
cam-front       & Front cam. img.    & .png      & 4.5GB         & network input, visualization           \\
cam-left        & Left cam. img.     & .png      & 4.5GB         & network input, visualization           \\
cam-right       & Right cam. img.    & .png      & 4.5GB         & network input, visualization           \\
cam-rear        & Rear cam. img.     & .png      & 4.5GB         & network input, visualization           \\
cam-dash        & Dash cam. img.     & .mp4      & 75MB          & reference video of annotation          \\
info\_calib       & Calibration values & .txt      & -             & calibration of 4DRT and LPC            \\
info\_condition  & Conditions         & .txt      & -             & conditional evaluation                 \\
info\_label      & Labels             & .txt      & 0.5MB         & training, evaluation                  \\ \hline\hline
\end{tabular}
\end{center}
}
\end{table}

\section{Details of the sensor suite}

We use waterproofed sensors with grade IP66 or higher, as mentioned in Section 3.1, for safe data collection in adverse weather conditions. 
Table \ref{tab:app_suite} summarizes the detailed information (i.e., model name, output data format, resolution, maximum operating distance, field of view (FOV), frames per second (FPS)) of the sensors that we install for the K-Radar data collection.

\begin{table}[h!]
{
\tabcounterint
\begin{center}
\caption{Sensor suite details: `Azi.', `Ele.', `res.' and `CEP' denotes azimuth, elevation angle, resolution, and circular error probability, respectively.}
\label{tab:app_suite}
\begin{tabular}{c|cccccc} \hline\hline
sensors                                                         & \begin{tabular}[c]{@{}c@{}}model\\ name\end{tabular}       & \begin{tabular}[c]{@{}c@{}}output\\ data\end{tabular}                          & resolution                                                     & \begin{tabular}[c]{@{}c@{}}max\\ range\end{tabular} & \begin{tabular}[c]{@{}c@{}}FOV\\ (Azi., Ele.)\end{tabular} & FPS \\ \hline
4D Radar                                                        & \begin{tabular}[c]{@{}c@{}}RETINA\\ -4ST\end{tabular}      & \begin{tabular}[c]{@{}c@{}}64$\times$256$\times$107$^{\circ}$$\times$37$^{\circ}$\\ size 4D tensor\end{tabular}         & \begin{tabular}[c]{@{}c@{}}0.06m/s,\\ 0.46m,1$^{\circ}$,1$^{\circ}$\end{tabular} & 118m                                                & \begin{tabular}[c]{@{}c@{}}107$^{\circ}$,\\ 37$^{\circ}$\end{tabular}          & 10  \\
\begin{tabular}[c]{@{}c@{}}long range\\ Lidar\end{tabular}      & os2-64                                                     & 131,072 3D points                                                              & \begin{tabular}[c]{@{}c@{}}0.1cm,\\ 0.18$^{\circ}$,0.35$^{\circ}$\end{tabular}    & 240m                                                & \begin{tabular}[c]{@{}c@{}}360$^{\circ}$,\\ 22.5$^{\circ}$\end{tabular}        & 10  \\
\begin{tabular}[c]{@{}c@{}}high res.\\ Lidar\end{tabular} & os1-128                                                    & 262,144 3D points                                                              & \begin{tabular}[c]{@{}c@{}}0.1cm,\\ 0.18$^{\circ}$,0.35$^{\circ}$\end{tabular}    & 120m                                                & \begin{tabular}[c]{@{}c@{}}360$^{\circ}$,\\ 45$^{\circ}$\end{tabular}          & 10  \\
\begin{tabular}[c]{@{}c@{}}4 stereo\\ cameras\end{tabular}      & ZED2i                                                      & \begin{tabular}[c]{@{}c@{}}8 1280$\times$720 size\\ images (left, right)\end{tabular} & \begin{tabular}[c]{@{}c@{}}1280x720\\ pixels\end{tabular}      & n/a                                                 & \begin{tabular}[c]{@{}c@{}}110$^{\circ}$,\\ 70$^{\circ}$\end{tabular}          & 30  \\
RTK-GPS                                                         & \begin{tabular}[c]{@{}c@{}}GPS500,\\ C94-M8P3\end{tabular} & \begin{tabular}[c]{@{}c@{}}latitude, longitude,\\ altitude\end{tabular}        & \begin{tabular}[c]{@{}c@{}}0.025m +\\ 1ppm CEP\end{tabular}    & n/a                                                 & n/a                                                        & 1   \\
2 IMUs                                                          & \begin{tabular}[c]{@{}c@{}}built-in\\ Lidar\end{tabular}   & 6-axis IMU data                                                                & n/a                                                            & n/a                                                 & n/a                                                        & 100                     \\ \hline\hline
\end{tabular}
\end{center}
}
\end{table}

\section{Criteria for weather conditions, road structures, and collecting time}

We establish conditions for each sequence according to the criteria in Table \ref{tab:app_criteria}, as mentioned in Section 3.2.

\begin{table}[h!]
{
\tabcounterint
\begin{center}
\caption{Detailed criteria for each condition.}
\label{tab:app_criteria}
\begin{tabular}{ccl} \hline\hline
Criteria                                                     & Name                                                   & \multicolumn{1}{c}{Detailed criteria}                                                                                              \\ \hline
                                                             & urban                                                  & \begin{tabular}[c]{@{}l@{}}{Roads} with four or more lanes and traffic lights, and ego-vehicle\\ average speed is around 60 km/h\end{tabular} \\
                                                             & highway                                                & \begin{tabular}[c]{@{}l@{}}{Roads} without traffic lights and ego-vehicle average speed\\ is around 100km/h\end{tabular}               \\
    & alleyway                                               & {Roads} with two to four lanes and buildings nearby                                                                   \\
                                                \begin{tabular}[c]{@{}c@{}}road\\ structures\end{tabular}             & suburban                                               & \begin{tabular}[c]{@{}l@{}}Two- or four-lane {roads} with rice paddies, fields and mountains\\ around it\end{tabular}               \\
                                                             & university                                             & The inner roads of KAIST                                                                                                            \\
                                                             & mountain                                               & Sloped {roads} with two to four lanes in a countryside                                                                                   \\
                                                             & \begin{tabular}[c]{@{}c@{}}parking\\ lots\end{tabular} & {Areas} for stopping or parking with other vehicles around                                                                         \\
                                                             & shoulder                                               & Parking spaces by the side of the road                                                                                            \\ \hline
                                                             & normal                                                 & Clear weather that does not meet the six weather conditions below                                                                  \\
                                                             & overcast                                               & Sunless, cloudy weather                                                                                                            \\
\begin{tabular}[c]{@{}c@{}}weather\\ conditions\end{tabular} & fog                                                    & \begin{tabular}[c]{@{}l@{}}Weather in which distant objects are dimly visible due to\\ omni-directional fog\end{tabular}           \\
                                                             & rain                                                   & Rainy weather                                                                                                                      \\
                                                             & sleet                                                  & Precipitation that consists of both rain and snow\\
                                                             %Combined precipitation and snowfall in the weather                                                                                 \\
                                                             & light snow                                             & Snowfall within approximately 1 cm per hour                                                                                        \\
                                                             & heavy snow                                             & Snowfall that exceed 1 cm per hour                                                                                          \\ \hline
time                                                         & day                                                    & Approximately 6:00 $\sim$ 16:00                                                                                                         \\
zone                                                         & night                                                  & Approximately 20:00 $\sim$ 4:00                                                                                                        \\ \hline\hline
\end{tabular}
\end{center}
}
\end{table}

\section{Details of annotation and calibration}

\subsection{Details of annotation}

\begin{figure}[h!]
{
  \figcounterint
  \centering
  \includegraphics[width=0.7\columnwidth]{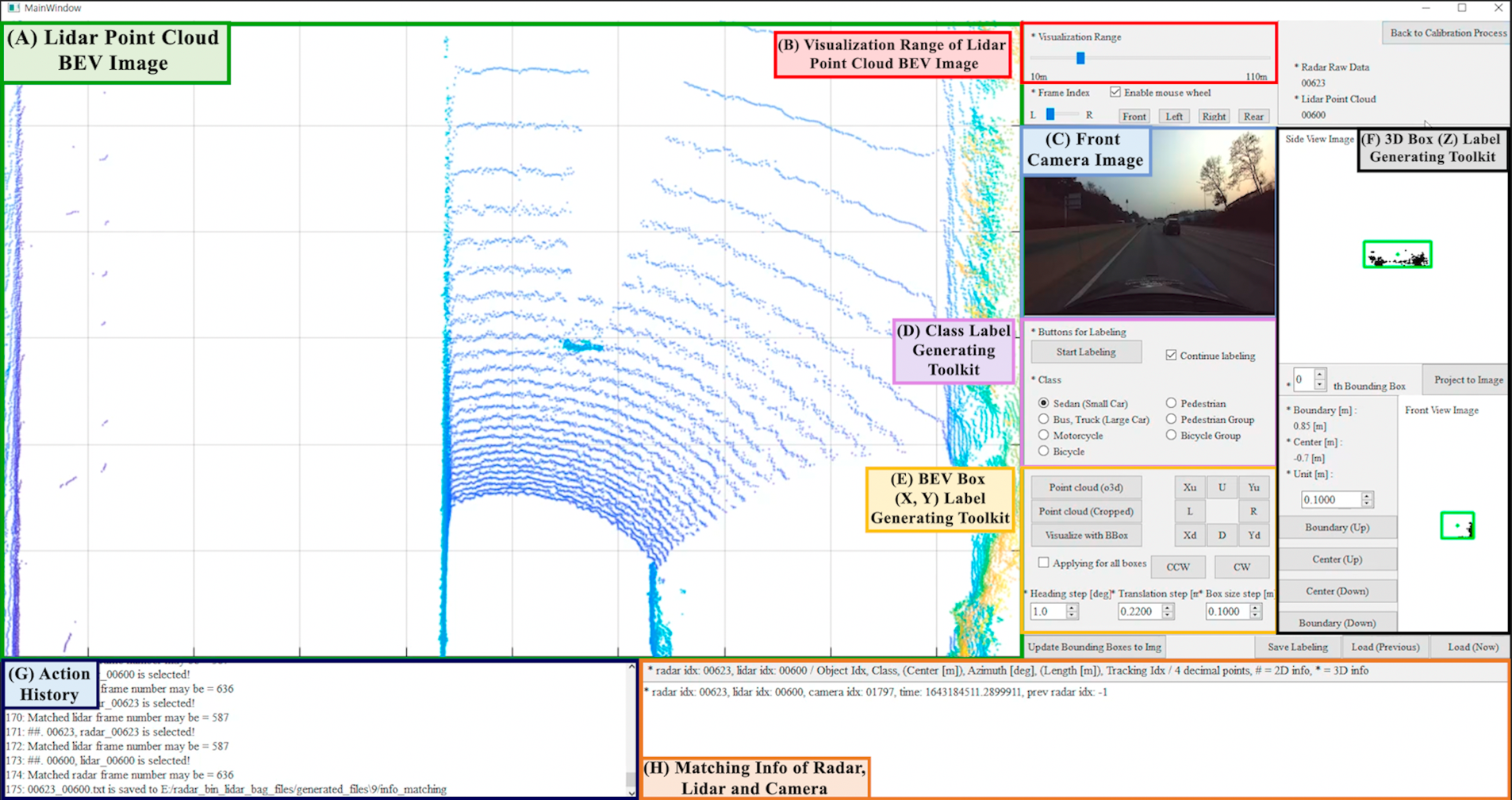}
  \caption{{A snippet} of the video clip that shows annotation process. (see Section J URL 3)}
  \label{annot_process}
}
\end{figure}

\textbf{Annotation process for calibrated LPC measurements}
As mentioned in Section 3.3, it is difficult to intuitively recognize the shape of objects in BEV-2D.
%As mentioned in Section 3.3, BEV-2D alone makes it difficult to intuitively check the shape of the object, which makes difficult to annotate the accurate 3D bounding box.
Therefore, we utilize the calibrated LPC (Section D.2) with a maximum calibration error of 0.5cm to enable accurate 3D bounding box annotations.
%Therefore, we annotate the 3D bounding box based on a LPC calibrated with an error of 0.5cm or less through the process to be described later in Section D.2.
We include the annotation program and code in the published devkits.
%We have also further expand the functions (e.g., annotating on BFS-2D) not provided in existing commercial programs by developing the own annotation program, which is all included in the devkits we provide.
The annotation program supports a resolution of 1.4 cm per pixel, resulting in a maximum annotation error of 0.7 cm.
%The annotation program we develop supports a resolution of 1.4 cm per pixel, resulting in the maximum annotation error to be 0.7 cm.
The annotation program can be used by following the two steps:
%The annotation program we provide can be used in two main processes as follows:
(1) annotate the BEV bounding box of an object in the visualized BEV LPC, (2) annotate the height and center point of the BEV bounding box.
%(1) After annotating the BEV bounding box for the visualized BEV LPC, (2) annotate the height and center point for the BEV bounding box.
In Figure \ref{annot_process}, we show the GUI and usage of the annotation program, and detailed instructions can be found in the video clip that is available at Section J URL 3.

\begin{figure}[h!]
{
  \figcounterint
  \centering
  \includegraphics[width=0.85\columnwidth]{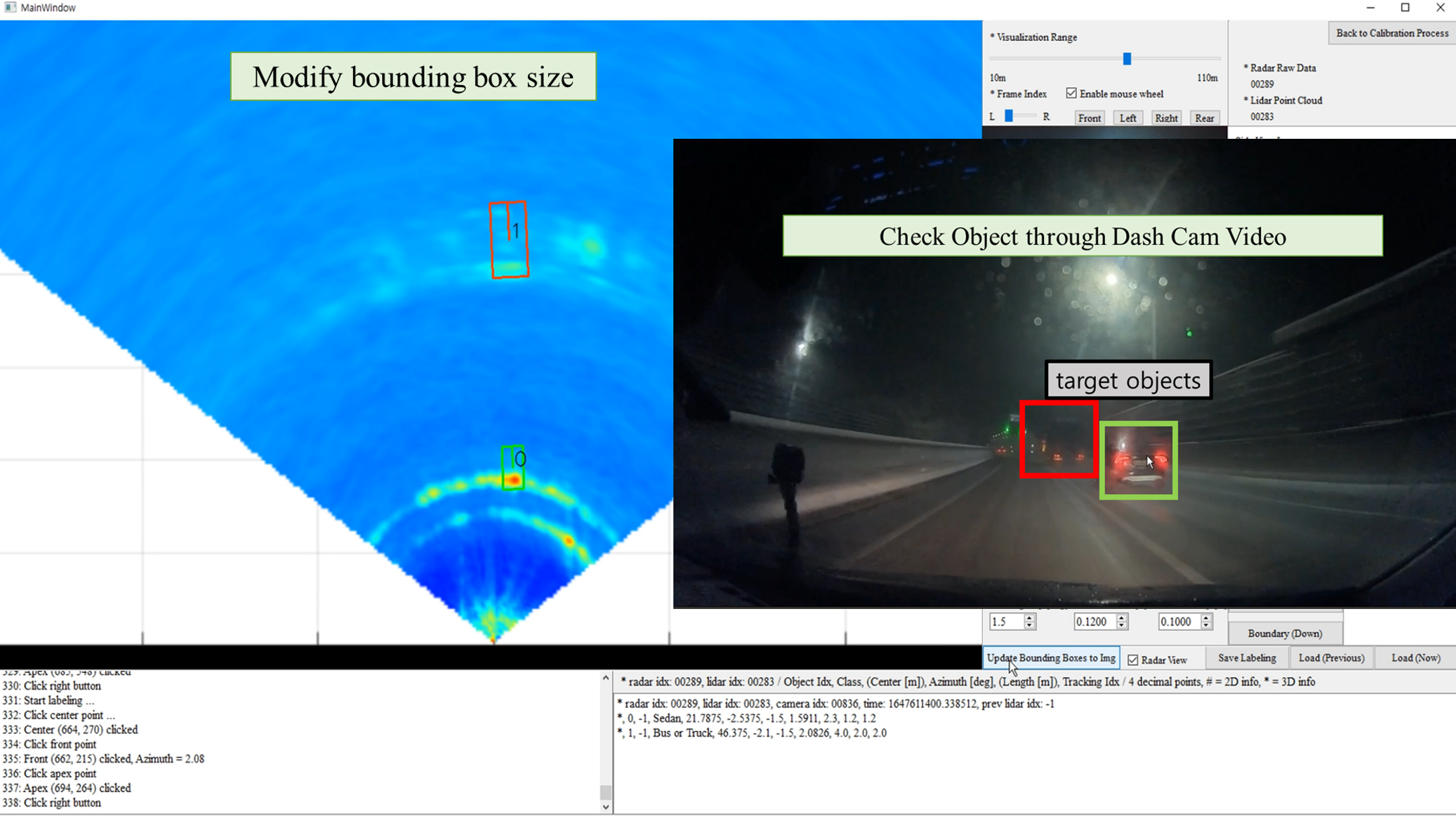}
  \caption{{A snippet} of the video clip that shows the annotation process in the absence of LPC measurements of objects. (see Section J URL 4)}
  \label{annot_process_bev2d}
}
\end{figure}

\textbf{Annotation process in the absence of LPC measurements of objects} 
As mentioned in Section 3.3, the annotation program we provide has a function to overlap the calibrated BEV-2D to the LPC so that annotations can be created even in the absence of LPC measurements of objects for various reasons such as adverse weather conditions.
%As mentioned in Section 3.3, the annotation program we provide has added the function to overlap the calibrated BEV-2D to the LPC, and is designed to be annotated even in the absence of LPC measurements of objects for various reasons such as adverse weather conditions.
%Since the shape of the object cannot be recognized based on the LPC in the absence of LPC measurements of the objects, annotation proceeds in the following order using BFS-2D of 4DRT.
To annotate objects in the absence of LPC measurements, the human annotator processes 3D bounding box annotation by referring to overlapped BEV-2D and dash camera images of the ego-vehicle.
The human annotator then verifies the height and size information of {the} 3D bounding box with BFS-2D, as shown in Figure 6-(b).
{We note that the height of the vehicle is set to a pre-defined value after checking the type of the vehicle through the dash cam image.}
Figure \ref{annot_process_bev2d} illustrates the GUI of the annotation program in the absence of LPC measurements, and more detailed instructions can be found in the video clip available at Section J URL 4.

\subsection{Details of the calibration between 4D Radar and Lidar}

\begin{figure}[h!]
{
  \figcounterint
  \centering
  \includegraphics[width=1.0\columnwidth]{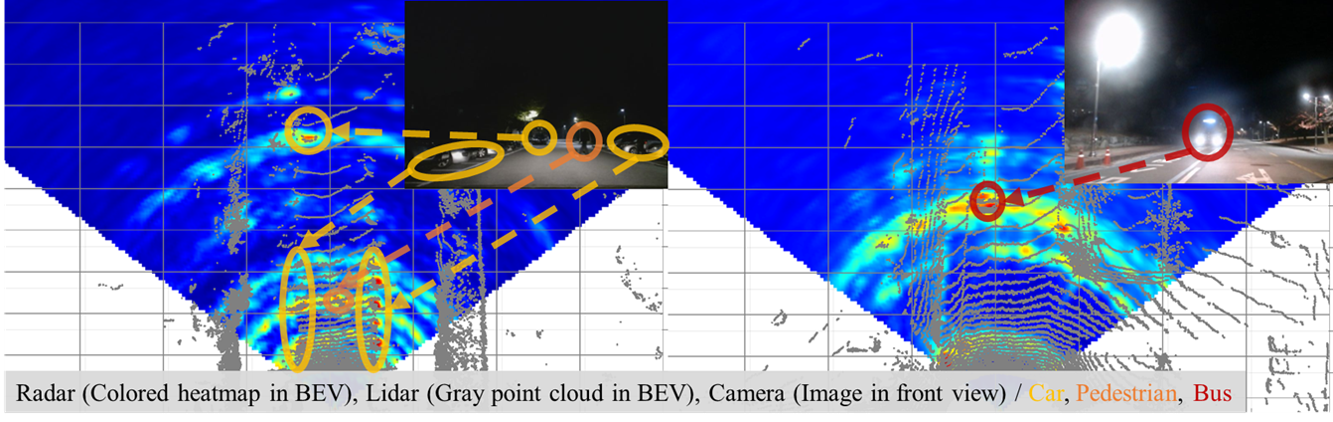}
  \caption{Examples of calibration clues.}
  \label{calibration_clue}
}
\end{figure}

\begin{figure}[h!]
{
  \figcounterint
  \centering
  \includegraphics[width=0.85\columnwidth]{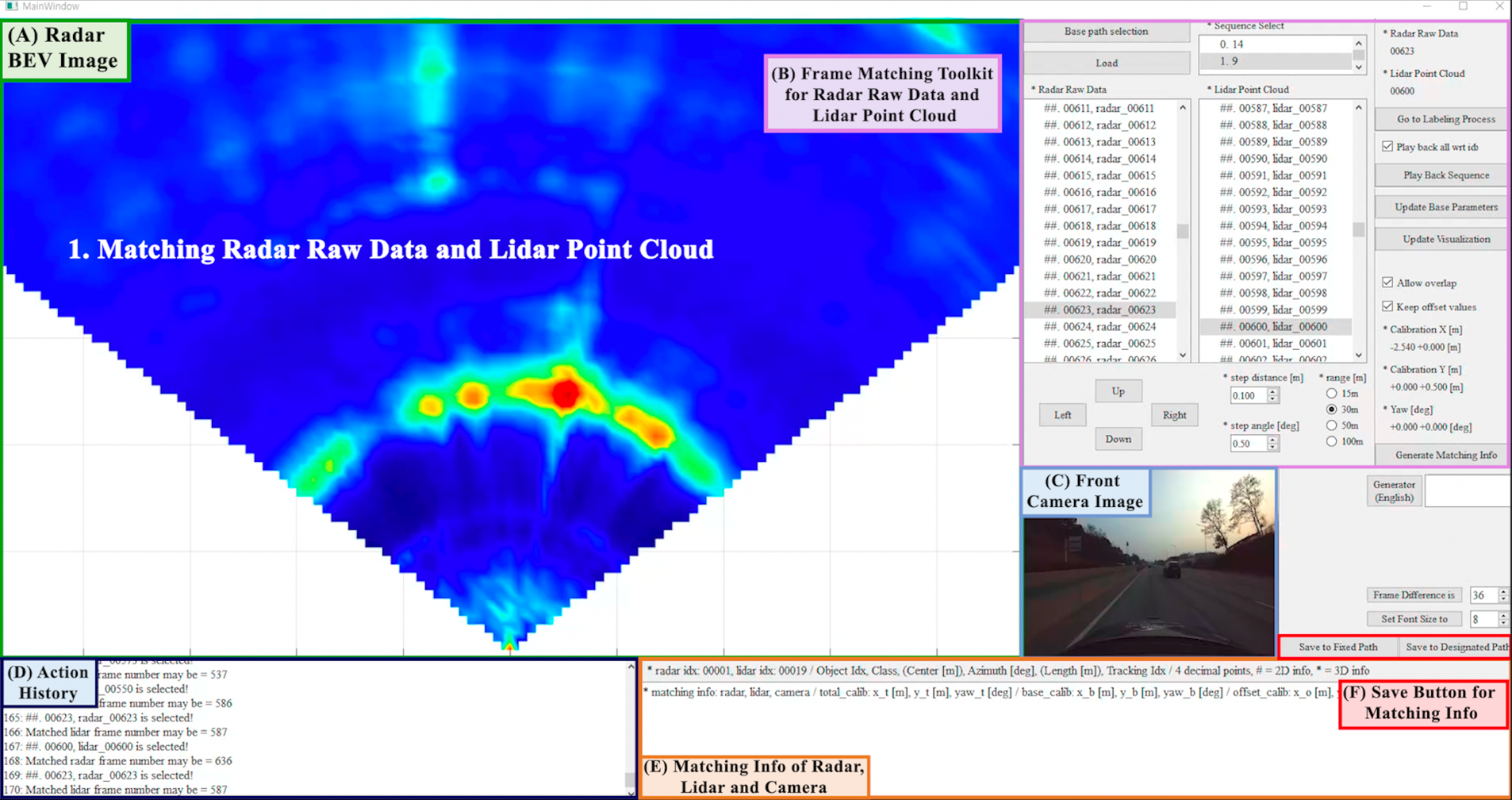}
  \caption{{A snippet} of the video clip that shows 4DRT/LPC calibration process through BEV-2D and LPC visualization. (see Section J URL 3)}
  \label{calibration_process}
}
\end{figure}

\begin{figure}[h!]
{
  \figcounterint
  \centering
  \includegraphics[width=1.0\columnwidth]{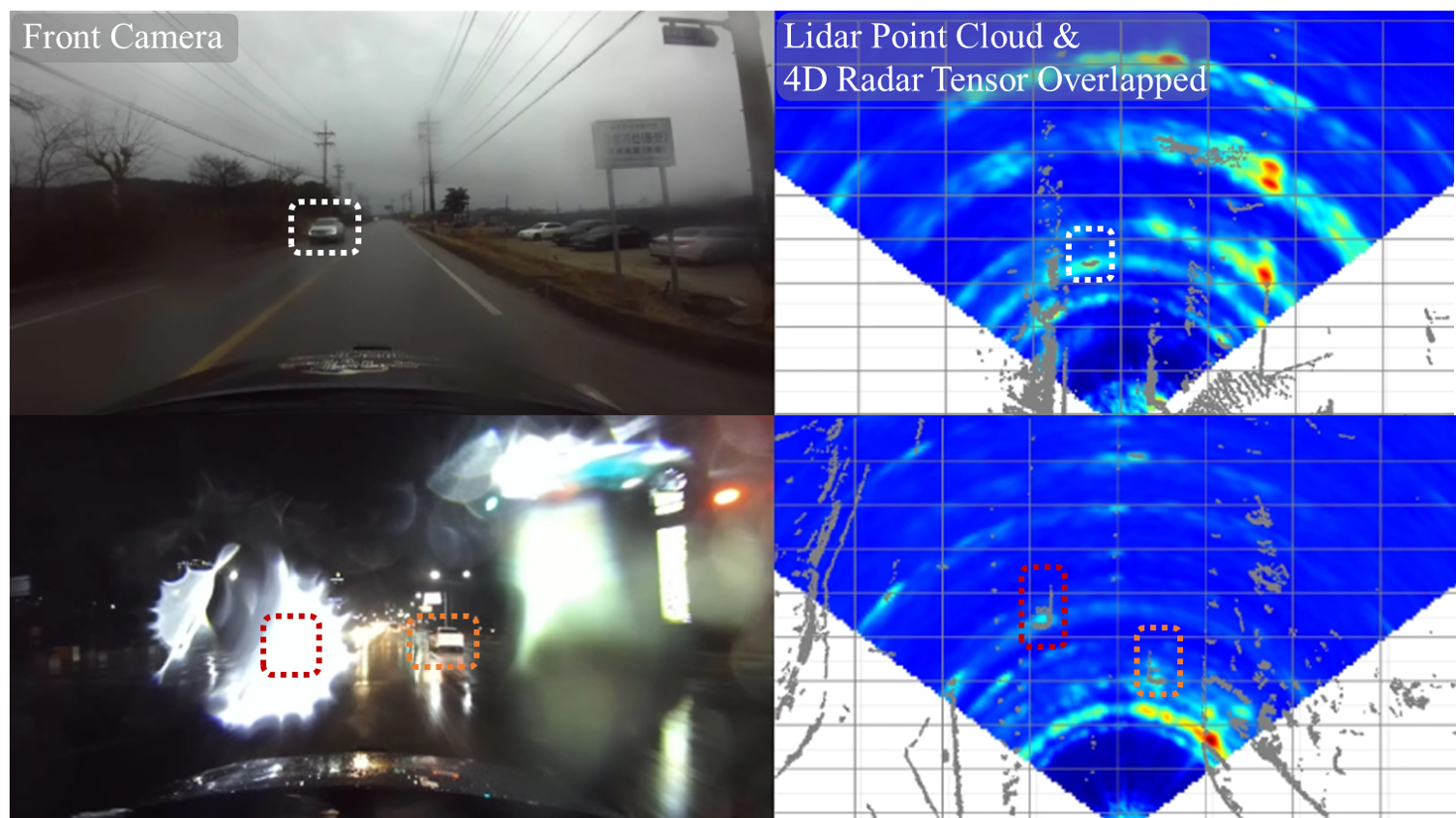}
  \caption{{A snippet} of the video clip that shows calibration results images for two different roads. (see Section J URL 5)}
  \label{calibration_result}
}
\end{figure}

Accurate calibration of the 4DRT and LPC is crucial to utilize the 3D bounding box annotated in the LPC as a label of the 4DRT.
%Accurate calibration of 4DRT and LPC is essential to utilize the 3D bounding box annnotated based on the LPC as a label of 4DRT. 
We utilize visualized BEV-2D and LPC as well as spatial information (sensor placement location) of all sensors to precisely calibrate 4DRT and LPC.
We develop a program that matches temporal offset (i.e., frame error) and spatial offset (i.e., 2D translation, yaw) through near-field (within about 30m) objects which are clearly visualized  (calibration clue shown in Figure \ref{calibration_clue}), as shown in Figure \ref{calibration_process}.
The GUI program in Figure \ref{calibration_process} supports a resolution of 1 cm per pixel, resulting in a maximum calibration error of 0.5 cm.
We have not considered the pitch angle difference between 4D Radar and Lidar, since we fix the sensors precisely perpendicular to the ground, resulting in no theoretical difference in the pitch angle.
We note that the video clip containing the calibration process (shown in Figure \ref{calibration_process}) is available in Section J URL 3, and the video clip containing the calibration result (shown in Figure \ref{calibration_result}) is available in Section J URL 5.
Through the calibration process, we note that the vehicle approaching from the other side matches correctly as shown in the calibration result (shown in Figure \ref{calibration_result}).

\subsection{Details of the calibration between Lidar and camera}

\begin{figure}[h!]
{
  \figcounterint
  \centering
  \includegraphics[width=0.5\columnwidth]{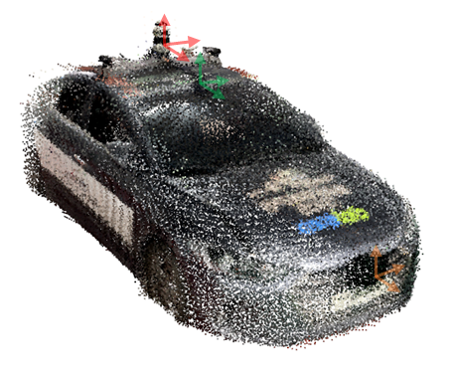}
  \caption{{The scanned 3D model of the ego-vehicle with sensor suite.}}
  \label{scanned_vehicle}
}
\end{figure}

\begin{figure}[h!]
{
  \figcounterint
  \centering
  \includegraphics[width=1.0\columnwidth]{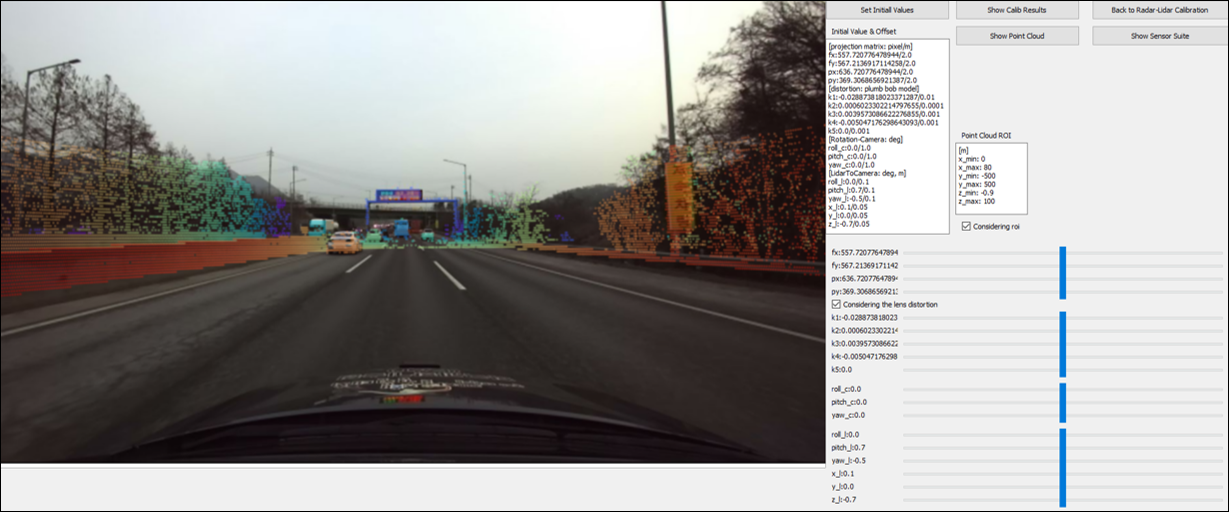}
  \caption{{The GUI program to fine-tune the calibration parameters between Lidar and camera.}}
  \label{lidar_camera_calibration_gui}
}
\end{figure}

\begin{figure}[h!]
{
  \figcounterint
  \centering
  \includegraphics[width=1.0\columnwidth]{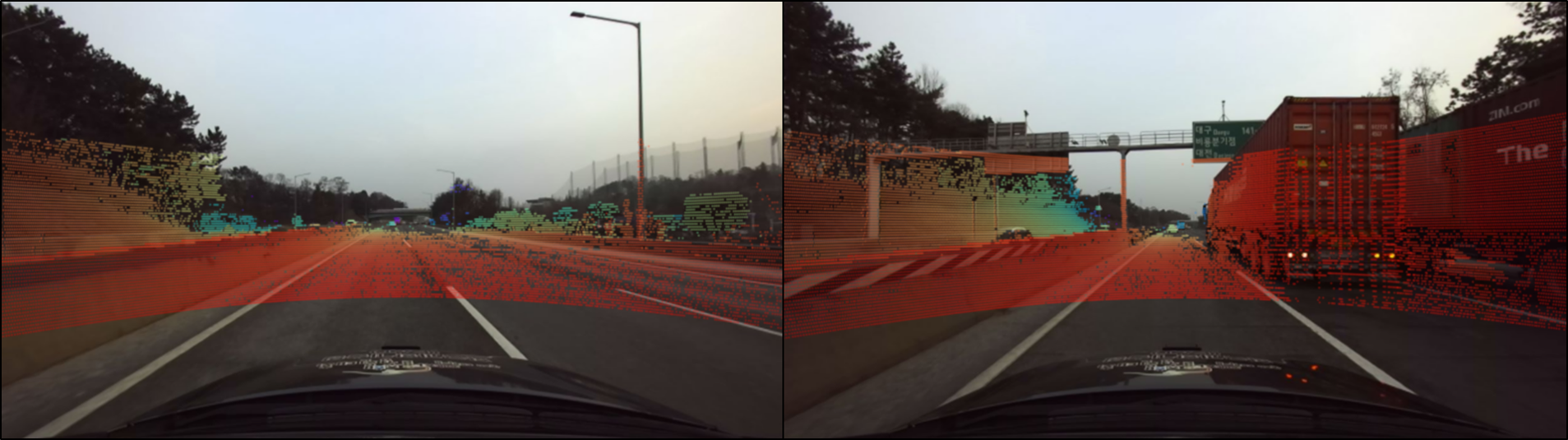}
  \caption{{Examples of calibration result between the camera and Lidar, where colored points on the front images show the projected points of the corresponding LPCs.}}
  \label{lidar_camera_calibration_result}
}
\end{figure}

{The calibration of the Lidar and camera is to determine a total of three parameters: 1) extrinsic parameters to define the relative position of the Lidar coordinate system (i.e., reference frame) and the camera coordinate system, 2) lens distortion parameters to correct camera distortion, and 3) intrinsic parameters to match each pixel in the pixel coordinate system with the points in the camera coordinate system. As shown in Figure} \ref{scanned_vehicle}, we {scan a 3D model of the ego-vehicle with the sensor suite using a Lidar scanner provided in iPhone 12 Pro} \citep{iphone_lidar}. {We extract the coarse position of each sensor from the scanned 3D vehicle model. Second, we extract the lens distortion parameters and the intrinsic parameters of the camera using the camera calibration process provided by ROS} \citep{ros_noetic}. {The previous two processes extract approximate calibration parameters, which may include calibration errors. Therefore, we construct a GUI program that can modify each parameter finely, as shown in Figure} \ref{lidar_camera_calibration_gui}, and fine-tune the parameters so that the measurements of the camera and the LIDAR at the close and far objects match accurately, as shown in Figure \ref{lidar_camera_calibration_result}.

% \begin{wrapfigure}{L}{0.5\columnwidth}
% {
%   \figcounterint
%   \centering
%   \includegraphics[width=0.5\columnwidth]{figa2.png}
%   \caption{\hl{The scanned 3D model of the ego-vehicle with the sensor suite.}}
%   \label{scanned_vehicle}
% }
% \end{wrapfigure}

\begin{figure}[h!]
{
  \figcounterint
  \centering
  \includegraphics[width=0.55\columnwidth]{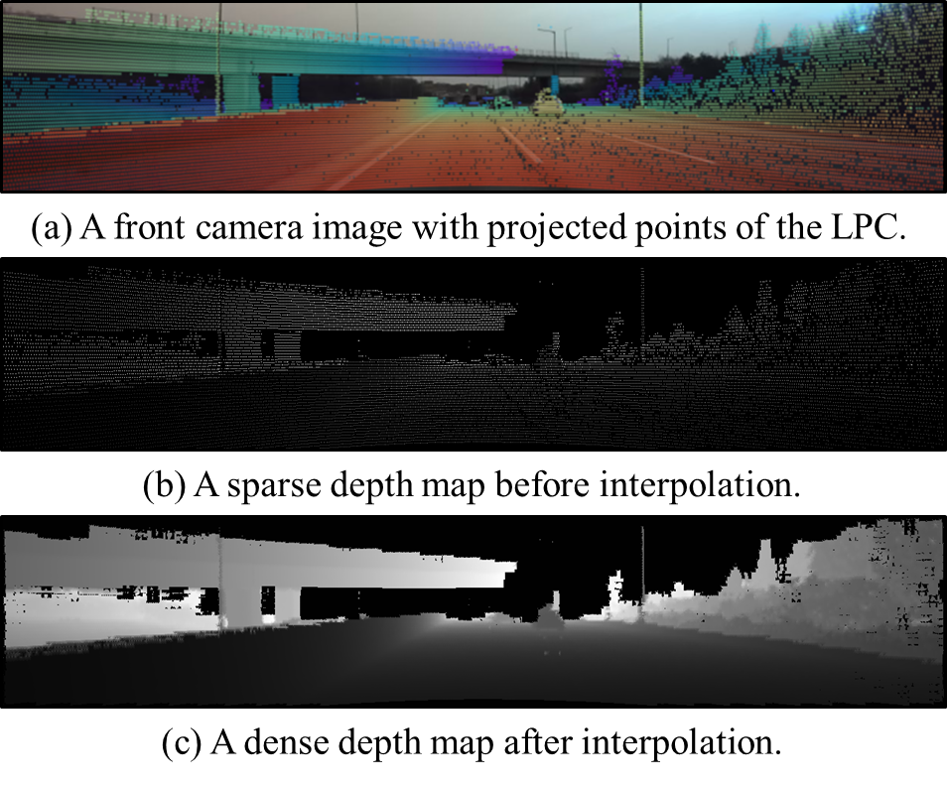}
  \caption{{An example of generating a depth map based on calibration result.}}
  \label{depth_map}
}
\end{figure}

In addition, we obtain the ground-truth depth value of the corresponding pixel through the points projected onto the camera image. Because the LPC is sparse, as shown in Figure \ref{depth_map}-(b), the dense depth map is provided through interpolation, as shown in Figure \ref{depth_map}-(c). We note that these depth maps can magnify the utilization of our dataset for the depth estimation tasks \citep{depth_estimation}, which is one of the most widely studied fields in computer vision.

\section{Details of baseline NNs}

In this section, we describe the common structures of RTNH and RTN, neck, and head of the baseline NNs, and {the structures of} 3D-SCB and 2D-DCB, which are the backbone of RTNH and RTN, respectively.

\subsection{Neck and head}
\begin{wrapfigure}{L}{0.5\columnwidth}
{
  \figcounterint
  \centering
  \includegraphics[width=0.5\columnwidth]{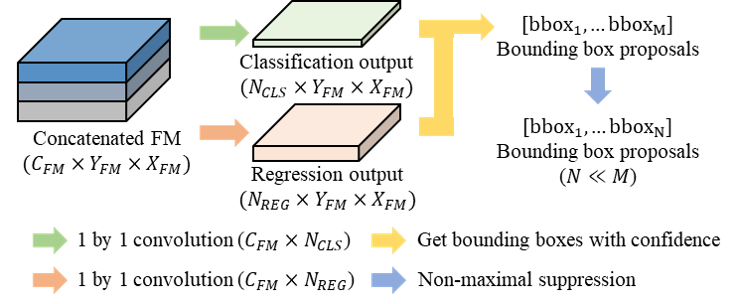}
  \caption{{Head structure.}}
  \label{head_structure}
}
\end{wrapfigure}

As mentioned in Section 3.4, both RTNH and RTN extract multiple feature maps (FMs) of different resolutions. 
Neck transforms the FMs into the same size by applying TransposeConv2D {and concatentes the transformed FMs} \citep{fpn}. %and \hl{to output a concatenated FM} \citep{fpn}. 
The size of the concatenated FM is ${C_{FM}}\times{Y_{FM}}\times{X_{FM}}$. $C_{FM}$, $Y_{FM}$, and $X_{FM}$ represent the number of channels of the concatenated FM, the number of grids for the left and right widths, and the number of grids for the front distance, respectively.
The head predicts the bounding boxes from the concatenated FM using an anchor-based method as in \citet{faster_rcnn}, and its structure is shown in Figure \ref{head_structure}.

We apply 1 by 1 convolution to the concatenated FM to extract classification and regression output for each grid, as shown in Figure \ref{head_structure}.
We use two anchor boxes with yaw angles of 0$^{\circ}$ and 90$^{\circ}$ for each class, resulting in $N_{CLS}=2$(Anchor)$+1$(Background)$=3$. 
In addition, we assign a total of eight parameters for each anchor: center point ($x_c,y_c,z_c$), length, width, height ($x_l,y_l,z_l$), $cos(yaw)$, and $sin(yaw)$ \citep{complex_yolo} of the bounding box, resulting in $N_{REG}=8×(N_{CLS}-1)=16$. 
We then extract $M$ bounding box proposals from the classification and regression outputs. 
In training process, proposals with an intersection over union (IOU) of 0.5 or more with respect to the ground-truth are classified as positive bounding boxes, and proposals with an IoU of less than 0.2 are classified as negative bounding boxes. 
We apply the focal loss \citep{focal} to cope with the problem of class imbalance between positive bounding boxes and negative bounding boxes, and apply the smooth L1 loss between the regression value and the target value.
During inference, an index with the largest logit value from the classification output is inferred as the proposal's class, and a confidence threshold of 0.3 is applied, so that low-confidence predictions are regarded as backgrounds. 
Thereafter, non-maximal suppression is applied to remove overlapping bounding boxes and finally a total of $N$ bounding boxes are obtained.

\subsection{3D-SCB}

As mentioned in Section 3.4, we extract FMs using 3D sparse conv blocks to reduce the usage of GPU memory, while still using height information from the 4DRT.
A 3D sparse conv block consists of a total of three consecutive 3D convolution layers.
We set the first 3D convolution layer as 3D sparse convolution layer \citep{sparse_conv} and the remaining 3D convolution layer as 3D submanifold convolution layer \citep{sub_sparse}.
The output of the 3D sparse conv block is a sparse FM with four dimensions (channel, height, width, length) of different resolutions.
% Each sparse FM is transformed into its dense tensor counterpart, and then {TransposeConv2D} is applied to the three-dimensional dense FMs, whose Z-axis is concatenated with the channel axis \citep{voxelrcnn}, resulting in dense FMs represented in BEVs with height information encoded.
Each sparse FM is transformed into its dense tensor counterpart, and then {TransposeConv2D} is applied to the three-dimensional dense FMs, resulting in dense FMs represented in BEVs with height information encoded.
Finally, all dense FMs are concatenated to produce the final concatenated FM, which is the input of the head.

\subsection{2D-DCB}
We construct a 2D dense conv backbone (2D-DCB) with 2D conv blocks, as mentioned in Section 3.4, to extract FMs without encoding the height information.
We utilize ResNet50 \citep{resnet} and ResNext101 \citep{resnext} as the 2D conv blocks whose performance has been validated on tasks such as classification \citep{cls_with_resnext} and object detection \citep{obj_det_with_resnext}.
We compare object detection performance for two variations, as shown in Table \ref{tab:rtn_variants}, and in Section 4.2, we show the results of 2D-DCB-ResNext101, which has higher performance among the two, as the representative result of RTN.

\begin{table}[h!]
{
\tabcounterint
\begin{center}
\caption{Performance of two variants of RTN.}
\label{tab:rtn_variants}
\begin{tabular}{c|ccc}
\hline\hline
backbone          & ${AP}_{3D}$ [\%] & ${AP}_{BEV}$ [\%] & GPU RAM [MB] \\ \hline
2D-DCB-ResNext101 & \textbf{40.12}  & \textbf{50.67}  & 520          \\
2D-DCB-ResNet50   & 39.86  & 49.37  & \textbf{257}          \\ \hline\hline
\end{tabular}
\end{center}
}
\end{table}

\section{Qualitative results of RTNH and PointPillars with additional discussion in various conditions}

\begin{figure}[h!]
{
  \figcounterint
  \centering
  \includegraphics[width=1.0\columnwidth]{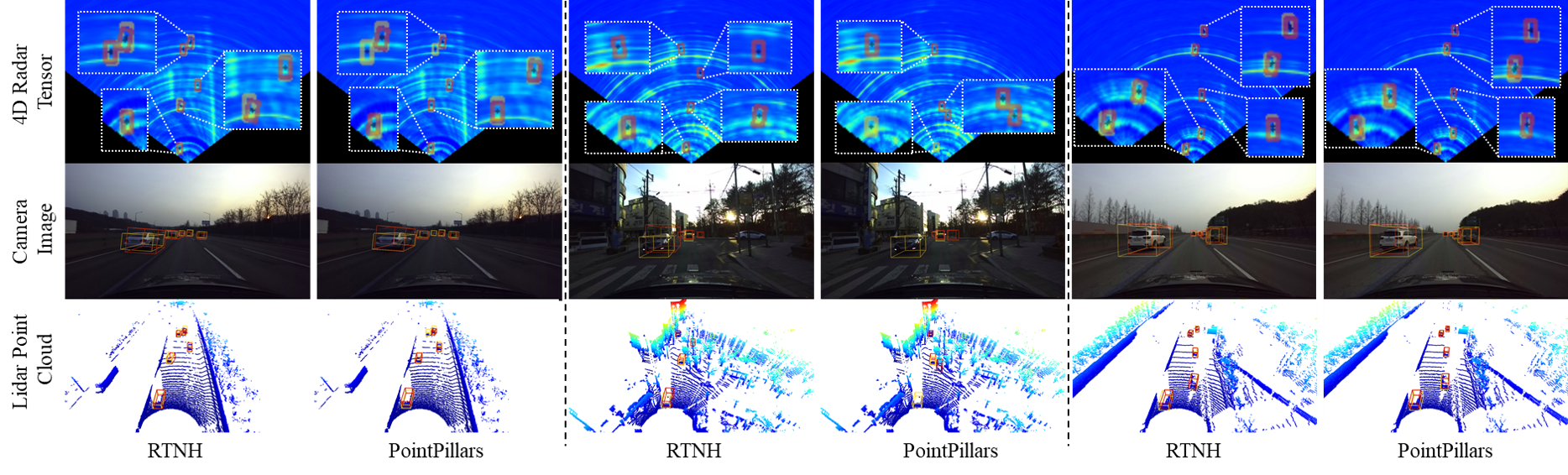}
  \caption{3D object detection results of RTNH (4DRT) and PointPillars (LPC) in a road environment {where} multiple vehicles exist. {We use yellow and red boxes to represent the ground truths and predictions, respectively.}}
  \label{qual1}
}
\end{figure}

\begin{figure}[h!]
{
  \figcounterint
  \centering
  \includegraphics[width=1.0\columnwidth]{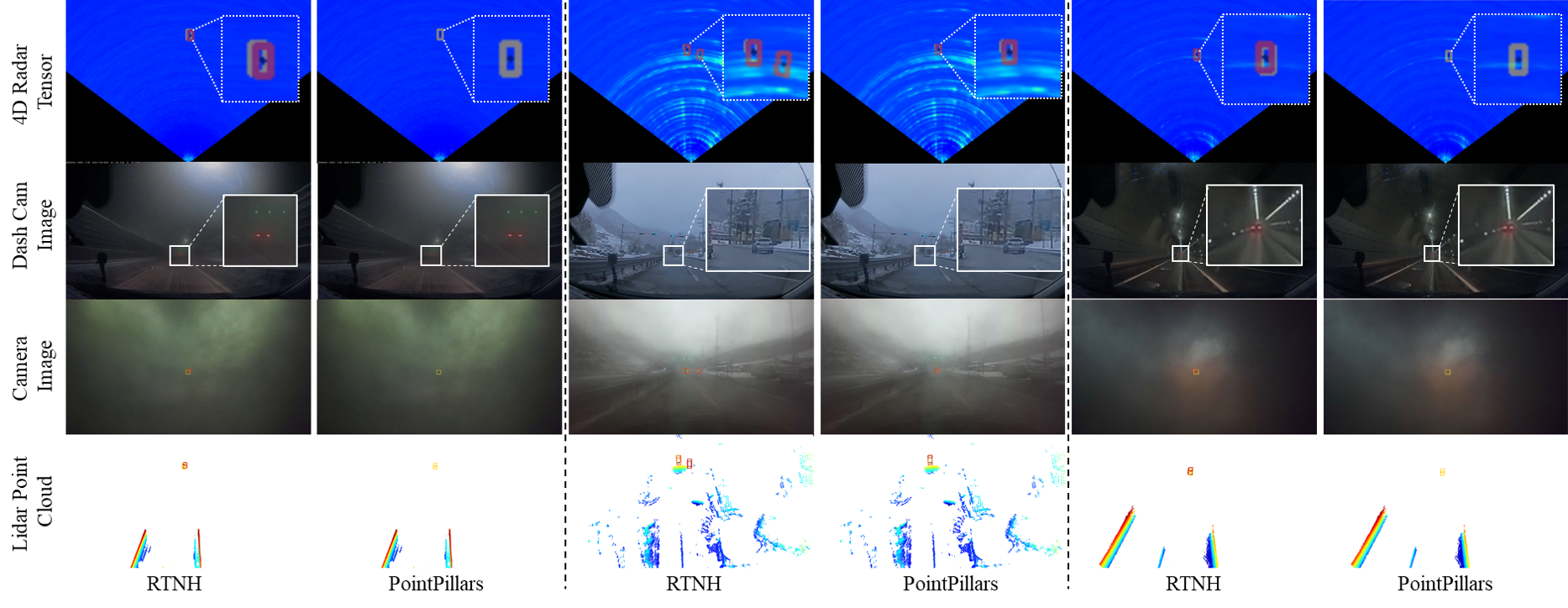}
  \caption{3D object detection results of RTNH (4DRT) and Point Pillars (LPC) {under} weather conditions with precipitation. From the left, sleet, light snow, and heavy snow. {We use yellow and red boxes to represent the ground truths and predictions, respectively.}}
  \label{qual2}
}
\end{figure}

We show the object detection results of RTNH and PointPillars \citep{pointpillars} under various weather conditions in Figure \ref{qual1} and \ref{qual2} with 3D bounding box labels, BEV-2D, front camera image, and LPC.
We also show images from the dash camera, since some of the outdoor camera measurements are unreliable due to the adverse weather conditions.

Figure \ref{qual1} shows the object detection results of RTNH and PointPillars for the road environments {where} multiple vehicles exist {under} weather conditions without precipitation (e.g., normal, overcast). 
As shown in Figure \ref{qual1}, RTNH produces similar or more robust detection results (robust to miss detection) compared to PointPillars.
The comparisons summarized in Table 4 and Figure \ref{qual1} show that 4D Radar has similar or more robust detection performance to Lidar in various road environments where multiple vehicles exist.
This indicates that 4D Radar can be sufficiently used alone as a perception sensor in autonomous driving.

{In addition, notice that the general AP for the normal condition can be lower than the overcast condition on K-Radar because of the following two reasons. One reason is that 4D Radar and Lidar are not affected by lighting condition, so that the detection performance of 4D Radar and Lidar for overcast condition cannot be lower than that for normal condition. Another reason is that the normal condition in K-Radar has more difficult situation than the overcast condition; the normal condition contains various situations including many vehicles parked along the side of alleyways, while the overcast condition in K-Radar does not have many vehicles on clear urban roads of two lanes, as shown in Table }\ref{tab:app_seq_dist}.

Figure \ref{qual2} shows the object detection results of RTNH and PointPillars under weather conditions with precipitation (e.g., sleet, light snow, heavy snow).
{As mentioned in Section A.1, Lidar can produce reliable LPC measurements in rain and light snow condition, but not in sleet and heavy snow conditions, since the sensor surfaces are covered by frost or snow.}
This can be seen in the LPC in Figure \ref{qual2}, and also by comparing the PointPillars results in Table 4.
% The reason why performance of RTNH in light snow is lower than that of PointPillars is explained as follows. The road environment, which collects data in light snow, is an environment with many iron structures, including guide rails in the center of the road. The iron structure is measured with high power by the 4D Radar. As a result, RTNH has a number of false alarms that incorrectly recognize iron structures as vehicles (i.e., false alarm), and as shown in Table 4, RTNH shows lower performance than PointPillars.

From the results of Figure \ref{qual2} and Table 4, we demonstrate that 4D Radar is a more robust sensor than Lidar in the adverse weathers. 
% Why do you need to say the next sentence 
We note that the inputs of RTNH and PointPillars are 4DRT and LPC, respectively. 
As mentioned in 4.3, we do not claim that RTNH is a better neural network architecture compared to PointPillars.
Instead, we demonstrate the robustness of 4D Radar in all weather conditions including adverse weathers.

\section{Consideration of the K-Radar dataset as a pre-training dataset for other Radar tensor datasets}

Pre-training on large scale datasets is well known to help neural networks converge faster \citep{imagenet_pre_train}. Therefore, we may consider using K-Radar as a pre-training dataset for other Radar tensor-based object detection datasets, or conversely, using other datasets as a pre-training dataset for K-Radar. 

However, we want to note that the pre-training on K-Radar does not directly guarantee a strong improvement on RADIATE. This is because the characteristics of K-Radar and RADIATE are inherently different.

First, the power measurements in K-Radar and RADIATE have different distributions due to the different type of Radars used. When a neural network is trained on a dataset and applied to process target data of different distribution, there will be a poorly degraded performance in the target domain, as we see in Lidar object detection networks trained and evaluated on different type of point clouds (e.g., Velodyne and Ouster) \citep{train_in_germany}.

Second, the resolution of RADIATE (0.175m) is higher than K-Radar (0.46m). This mismatch of resolution can also adversely affect the detection performance as usually seen in Lidar object detection networks trained on NuScenes (32-channels) and evaluated on KITTI (64-channels) \citep{train_in_germany}.

Third, the data distributions are significantly different. K-Radar data is collected in South Korea where cars drive on the right, while RADIATE is collected in the U.K where cars drive on the left.

The above reasons apply to other Radar tensor-based datasets as well as RADIATE. For these reasons, it is difficult to expect performance improvement by using K-Radar as a pre-training dataset for other Radar tensor-based datasets and vice versa.

\section{Performance of RTNH on wider areas and multiple classes}
We present the detection performance of RTNH over broader areas and multiple classes in Table \ref{app:multi_class}. The results in Table \ref{app:multi_class} differ from those in Table \ref{tab:lidar_radar} in two main aspects: Firstly, the evaluation extends from a narrow to a wider scope with dimensions of x, y, z in the ranges of 0 to 72, -16 to 16, and -2 to 7.6 meters, respectively. Secondly, the performance now encompasses the Bus or Truck class alongside the Sedan class, thereby representing both compact and large vehicle categories. Moreover, we indicate the performance of the Bus or Truck class with a '-' for both rain and sleet conditions, as they are not present under these conditions.

\begin{table}[h!]
{
\tabcounterint
\begin{center}
\caption{Performance of RTNH on wider areas and multiple classes}
\label{app:multi_class}
\begin{tabular}{c|c|llllllll}
\hline\hline
Class & Metric & Total         & \begin{tabular}[c]{@{}l@{}}nor-\\ mal\end{tabular} & \begin{tabular}[c]{@{}l@{}}over-\\ cast\end{tabular} & fog           & rain          & sleet         & \begin{tabular}[c]{@{}l@{}}light\\ snow\end{tabular} & \begin{tabular}[c]{@{}l@{}}heavy\\ snow\end{tabular} \\ \hline
\multirow{2}{*}{Sedan} & $AP_{3D} [\%]$     & {48.2} & 45.5    & {58.8} & {79.3} & 40.3 & {48.1} & {65.6} & {52.6} \\
& $AP_{BEV} [\%]$ & {56.7} & {53.8}    & {68.3} & {89.6} & {49.3} & {55.6} & {69.4} & {60.3} \\ \hline
\multirow{2}{*}{Bus or Truck}& $AP_{3D} [\%]$ & 34.4 & {25.3} & 31.1 & - & {-} & 28.5 & 78.2 & 46.3 \\
& $AP_{BEV} [\%]$ & 45.3 & 31.8 & 32.0 & - & - & 34.4 & 89.3 & 78.0 \\ \hline\hline
\end{tabular}
\end{center}
}
\end{table}

\section{Details of devkits}

\begin{figure}[h!]
{
  \figcounterint
  \centering
  \includegraphics[width=0.9\columnwidth]{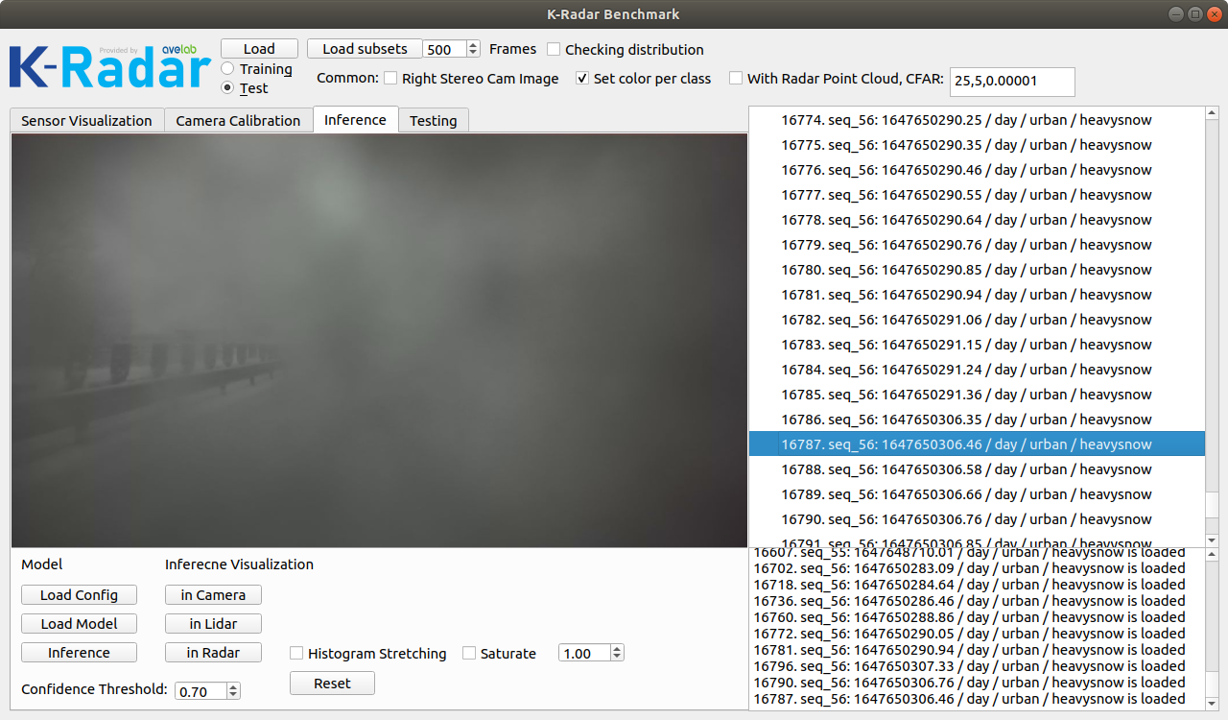}
  \caption{A snippet of the video clip that shows GUI-based program for visualization and neural network inference. (see Section J URL 6)}
  \label{devkits}
}
\end{figure}

To facilitate the experiments on various neural network structures, we provide modularized neural network training codes that can manage each experiment with a single configuration file. 
We also provide GUI-based programs for visualization and neural network inference, as shown in Figure \ref{devkits}, to facilitate inference on large amounts of data. 
We provide a video clip on how to use the program, which can be found through Section J URL 6, and all codes for devkits can be downloaded from Section J URL 1.

%%%%%%%%%%%%%%%%%%%%%%%%%%%% Until Here %%%%%%%%%%%%%%%%%%%%%%%%%%%%%%%%%%%%5

\section{Relevant URLs}

(1) Publication of datasets and complete devkits code (learning, evaluation, reasoning, visualization, labeling programs): \url{https://github.com/kaist-avelab/K-Radar}

(2) The video clip showing each sensor measurement dynamically changing during driving under the heavy snow condition: \url{https://www.youtube.com/watch?v=TZh5i2eLp1k&t=103s}

(3) The video clip showing the 4DRT/LPC calibration and annotation process: \url{https://www.youtube.com/watch?v=ylG0USHCBpU&t=152s}

(4) The video clip showing the annotation process in the absence of LPC measurements of objects: \url{https://www.youtube.com/watch?v=ILlBJJpm4_4&t=8s}

(5) The video clip showing calibration results: \url{https://www.youtube.com/watch?v=U4qkaMSJOds&t=10s}

(6) The video clip showing the GUI-based program for visualization and neural network inference: \url{https://www.youtube.com/watch?v=MrFPvO1ZjTY&t=3s}

(7) {The video clip showing the information regarding tracking for multiple objects on the roads:} \url{https://www.youtube.com/watch?v=8mqxf58_ZAk}

{
\small
\bibliography{neurips_data_2022}
}

\end{document}